\documentclass[12pt]{article}
 \usepackage[margin=0.99in]{geometry}

\newcommand{\remind}[1]{{\color{blue}#1}}
\newcommand{\issue}[1]{{\color{red}#1}}

\usepackage{kz_style} 

\usepackage{mathrsfs}
\usepackage{comment}

\makeatletter
\def\@fnsymbol#1{\ensuremath{\ifcase#1\or  \natural \or \dagger\or * \or \ddagger\or
   \mathsection\or \mathparagraph\or \|\or **\or \dagger\dagger
   \or \ddagger\ddagger \else\@ctrerr\fi}}
\makeatother

\title{\LARGE Multi-Agent Reinforcement Learning: A Selective Overview of  Theories and  Algorithms}
 
\begin{document} 
\author{Kaiqing Zhang\thanks{Department of Electrical and Computer Engineering \&  Coordinated Science Laboratory, University of Illinois at Urbana-Champaign, 1308 West Main, Urbana, IL,  61801, USA. Email: \{kzhang66, basar1\}@illinois.edu. Writing of this chapter was supported in part by the US Army Research Laboratory (ARL) Cooperative Agreement W911NF-17-2-0196, and in part by the Air Force Office of Scientific Research (AFOSR) Grant FA9550-19-1-0353.}  
 \and Zhuoran Yang\thanks{Department of Operations Research and Financial Engineering, Princeton University, 98 Charlton St, Princeton, NJ, 08540, USA.  {Email: zy6@princeton.edu}.} \and Tamer Ba\c{s}ar$^\natural$}
\date{}
%
%
\maketitle

\begin{abstract}
Recent years have witnessed significant advances in reinforcement learning (RL), which has registered tremendous success in solving various sequential decision-making problems in machine learning. Most of the successful RL applications, e.g., the games of Go and Poker, robotics, and autonomous driving, involve the participation of more than one single agent, which naturally  fall into the realm of multi-agent RL (MARL), a domain with a relatively long history, and   has recently re-emerged due to advances in single-agent RL techniques. Though empirically successful, theoretical foundations for MARL are relatively lacking in the literature. In this chapter, we provide a selective overview of MARL, with focus on  algorithms backed by theoretical analysis. More specifically, we review the theoretical results of MARL algorithms mainly within two representative frameworks, Markov/stochastic games and extensive-form games, in accordance with the types of tasks they address, i.e., fully cooperative, fully competitive, and a  mix of the two. We also introduce  several significant but challenging applications of these  algorithms. Orthogonal to the existing reviews on MARL,  we highlight several new angles and taxonomies of MARL theory, including learning in extensive-form games, decentralized MARL with networked agents, MARL in the mean-field regime, (non-)convergence of policy-based methods for learning in games, etc. Some of the new angles extrapolate from our own research endeavors and interests. 
 Our overall goal with this chapter is, beyond  providing an assessment of the current state of the field on the mark, to identify fruitful future research directions on   theoretical studies of MARL.  We   expect this chapter to serve as continuing stimulus for researchers interested in  working on this exciting while challenging topic.  
\end{abstract}

\section{Introduction}\label{sec:introduction}

Recent years have witnessed sensational advances of   reinforcement learning (RL) in many prominent  sequential  decision-making problems, such as playing the game of Go \cite{silver2016mastering, silver2017mastering}, playing   real-time strategy games  \cite{OpenAI_dota,alphastarblog},   robotic control  \cite{kober2013reinforcement,lillicrap2016continuous}, playing card games \cite{brown2017libratus,brown2019superhuman}, and autonomous driving \cite{shalev2016safe}, especially 
accompanied with  the development of  deep neural networks (DNNs) for function approximation \cite{mnih2015human}.  
Intriguingly, most of the successful applications involve the participation of more than one single agent/player\footnote{Hereafter, we will use \emph{agent} and \emph{player} interchangeably. }, which should be modeled systematically as multi-agent RL (MARL) problems. Specifically, MARL  addresses the sequential decision-making problem of  multiple autonomous  agents that operate in a common environment, each of which aims to optimize its own long-term return by interacting with the environment and other agents \cite{bu2008comprehensive}.  Besides the aforementioned popular ones, learning in multi-agent systems finds potential  applications in other subareas, including cyber-physical systems \cite{adler2002cooperative,wang2016towards}, finance  \cite{lee2002stock,lee2007multiagent}, sensor/communication networks \cite{cortes2004coverage,choi2009distributed}, and social science \cite{castelfranchi2001theory,leibo2017multi}.

Largely, MARL algorithms can be placed   into   three groups, \emph{fully cooperative}, \emph{fully competitive}, and \emph{a mix of the two},  depending on the types of settings they address. In particular, in  the cooperative setting, agents   collaborate to optimize a common long-term return; while in the competitive setting, the return of agents usually sum up to zero. The mixed setting involves both cooperative and competitive agents, with general-sum returns. Modeling disparate MARL settings requires frameworks spanning from optimization theory, dynamic programming, game theory, and decentralized control, see \S\ref{subsec:MARL_framework} for more detailed discussions. In spite of these   existing multiple frameworks, several challenges in MARL  are in fact common across the different  settings, especially for the theoretical analysis.  Specifically, first, the learning goals in MARL are \emph{multi-dimensional}, as the objectives of all agents are not necessarily aligned, which brings up the challenge of dealing with equilibrium points, as well as  some additional   performance criteria beyond return-optimization, such as  the efficiency of communication/coordination, and  robustness against potential adversarial agents. Moreover, as all agents are improving their policies according to their own interests   concurrently, the environment faced by each agent becomes \emph{non-stationary}. This breaks or invalidates the basic framework of most  theoretical analyses in the single-agent setting. Furthermore, the joint action space that increases exponentially with the number of agents may cause scalability issues, known as the \emph{combinatorial nature} of MARL \cite{hernandez2018multiagent}. Additionally, the information structure, i.e., \emph{who knows what},  in MARL is more involved, as each agent has limited access to the observations of others, leading to possibly suboptimal decision rules locally. A detailed elaboration on the underlying  challenges can be found in \S\ref{sec:challenges}.

There has in fact  been no shortage of efforts attempting to address the above   challenges. See \cite{bu2008comprehensive} for a comprehensive overview of earlier theories and algorithms on MARL. Recently, this  domain has gained resurgence of interest due to the advances of single-agent RL techniques. Indeed, a huge volume of work on MARL has appeared   lately, focusing on either identifying new learning criteria and/or setups \cite{foerster2016learning,zazo2016dynamic,zhang2018fully,subramanian2019reinforcement}, or developing new algorithms for existing setups, thanks to the development of deep learning \cite{heinrich2016deep,lowe2017multi,foerster2017counterfactual,gupta2017cooperative,omidshafiei2017deep,kawamura2017neural,zhang2019monte}, operations  research \cite{mazumdar2018convergence,jin2019minmax,zhang2019policyb,sidford2019solving}, and multi-agent systems \cite{oliehoek2016concise,arslan2017decentralized,yongacoglu2019learning,zhang2019online}.  
Nevertheless, not all the efforts are placed under rigorous theoretical footings, partly due to the limited understanding of even single-agent deep RL theories, and partly due to the inherent challenges in  multi-agent settings. As a consequence, it is imperative to review and organize the MARL algorithms with theoretical guarantees, in order to highlight the boundary of existing research endeavors, and stimulate potential future directions on this topic.   
 
In this chapter, we provide a selective overview of  theories and algorithms in MARL, together with several significant while challenging applications. More specifically, we focus on two representative frameworks of MARL, namely, Markov/stochastic games and extensive-form games, in discrete-time settings as in standard single-agent RL. In conformity with the aforementioned three groups, we review and pay particular attention to   MARL algorithms with convergence and complexity analysis, most of which are fairly recent. 
With this focus in mind, we note that our overview is by no means comprehensive. 
In fact, besides the classical reference  \cite{bu2008comprehensive}, 
there are several other reviews on MARL that have appeared recently, due to the resurgence of MARL \cite{hernandez2017survey,hernandez2018multiagent,nguyen2018deep,oroojlooyjadid2019review}. 
We would like to emphasize that these reviews provide views and taxonomies  that are complementary   to ours:  \cite{hernandez2017survey} surveys the works that are specifically devised to address \emph{opponent-induced non-stationarity}, one of the challenges we discuss in \S\ref{sec:challenges}; \cite{hernandez2018multiagent,nguyen2018deep} are relatively more comprehensive, but with the focal point on \emph{deep} MARL, a subarea with scarce theories thus far; \cite{oroojlooyjadid2019review}, on the other hand, focuses on  algorithms in the  \emph{cooperative} setting only, though the review within this setting is extensive. 

Finally, we spotlight several new angles and  taxonomies that are comparatively underexplored in the existing MARL reviews, primarily owing to our own research endeavors and interests. First, we discuss the framework of extensive-form  games in MARL, in addition to the conventional one of Markov games, or even simplified repeated games \cite{bu2008comprehensive,hernandez2017survey,hernandez2018multiagent}; second, we summarize the progresses of a recently boosting subarea: decentralized MARL with \emph{networked} agents, as an extrapolation of our early works on this \cite{zhang2018fully,zhang18cdc,zhang2018finite}; third, we bring about the \emph{mean-field} regime into MARL, as a remedy for the case with an extremely large  population of agents; fourth, we highlight some recent advances in optimization theory, which shed lights on the (non-)convergence of policy-based methods for MARL, especially zero-sum games. We have also reviewed some of the literature on MARL in partially observed settings, but without using deep RL as heuristic solutions.  
We expect these new angles to help identify fruitful future research directions, and more importantly, inspire researchers with interests in establishing   rigorous theoretical foundations  on MARL.   









\vspace{8pt}
{\noindent \bf Roadmap.}
\vspace{2pt}
The remainder of this chapter is organized as follows. In \S\ref{sec:background}, we introduce the background of MARL: standard algorithms for single-agent RL, and  the  frameworks of MARL. In \S\ref{sec:challenges}, we summarize several challenges in developing MARL theory, in addition to the single-agent counterparts. A series of MARL algorithms, mostly with theoretical guarantees, are reviewed and organized in \S\ref{sec:algorithms},  according to the types of tasks they address. In \S\ref{sec:applications}, we briefly introduce a few recent successes of MARL driven by the algorithms mentioned, followed by conclusions and several open  research directions outlined in \S\ref{sec:conclusion}.

\section{Background}\label{sec:background}

In this section,  we provide the necessary background on reinforcement learning, in both single- and multi-agent settings. 

\subsection{Single-Agent RL}
A reinforcement learning agent is  modeled to perform sequential decision-making by 
interacting with the environment. The environment is usually formulated as an infinite-horizon discounted Markov decision process (MDP), henceforth referred to as Markov decision process\footnote{Note that there are several other standard formulations of MDPs, e.g., time-average-reward setting and finite-horizon episodic setting. Here, we only present the classical infinite-horizon discounted setting for ease of exposition.}, which is formally defined as follows.

\begin{definition}
	A \emph{Markov decision process} is defined by a tuple $(\cS,\cA,\cP,R,\gamma)$, where $\cS$ and $\cA$ denote the state and action spaces, respectively;  $\cP:\cS\times\cA\to\Delta(\cS)$ denotes the transition probability from any state $s\in\cS$ to any state $s'\in\cS$ for any given action $a\in\cA$; $R:\cS\times\cA\times\cS\to\RR$ is the reward function that determines the immediate reward received by the agent for a transition from $(s,a)$ to $s'$; $\gamma\in[0,1)$ is the discount factor that trades off the instantaneous and future rewards. 
\end{definition}

As a standard model, MDP has been widely adopted to characterize the decision-making of an agent with \emph{full observability} of the system state $s$.\footnote{The partially observed MDP (POMDP) model is usually advocated when the agent has no access to the exact system state but only an \emph{observation} of the state. See {\cite{monahan1982state,cassandra1998exact}} for more details on the POMDP model.}  
At each time $t$, the agent chooses to execute an action $a_t$ in face of the system state $s_t$, which causes the system to transition to $s_{t+1}\sim \cP(\cdot\given s_t,a_t)$. Moreover, the agent receives an instantaneous  reward $R(s_t,a_t,s_{t+1})$. 
The goal of solving the MDP is thus to find a policy $\pi:\cS\to\Delta(\cA)$, a mapping from the state space $\cS$ to the distribution over the action space $\cA$, so that $a_t\sim \pi(\cdot\given s_t)$ and the discounted accumulated reward 
\$
\EE\bigg[\sum_{t\geq 0}\gamma^t R(s_t,a_t,s_{t+1})\bigggiven a_t\sim\pi(\cdot\given s_t),s_0\bigg]
\$ is maximized. 
Accordingly, one can define the \emph{action-value function} (Q-function) and \emph{state-value function} (V-function)   under policy $\pi$ as
\$
Q_\pi(s,a) & =\EE\bigg[\sum_{t\geq 0}\gamma^t R(s_t,a_t,s_{t+1})\bigggiven a_t\sim\pi(\cdot\given s_t),a_0=a,s_0=s\bigg], \\
 V_\pi(s) & =\EE\bigg[\sum_{t\geq 0}\gamma^t R(s_t,a_t,s_{t+1})\bigggiven a_t\sim\pi(\cdot\given s_t),s_0=s\bigg]
\$
for any $s\in\cS$ and $a\in\cA$, which are the discounted accumulated reward starting from $(s_0,a_0)=(s,a)$ and $s_0=s$, respectively. The ones corresponding to the optimal policy $\pi^*$ are usually referred to as the \emph{optimal Q-function} and the \emph{optimal state-value  function}, respectively. 

By virtue of the Markovian property, the  optimal policy can be obtained by dynamic-programming/backward induction approaches, e.g., value iteration and policy iteration algorithms \cite{bertsekas2005dynamic}, which require  the knowledge of the model,  i.e., the transition probability and the form of reward function. 
Reinforcement learning, on the other hand,  is to find such an optimal policy without knowing the model. The RL agent learns the policy from experiences collected by interacting with  the environment.  
By and large, RL algorithms can be categorized into two mainstream types, \emph{value-based} and \emph{policy-based} methods.

\subsubsection{Value-Based Methods} \label{eq:value-based-methods}
Value-based RL methods are devised to find a good estimate of the state-action value function, namely, the optimal Q-function $Q_{\pi^*}$. The (approximate) optimal policy can then be extracted by taking the greedy action of the Q-function estimate. One of the most popular value-based algorithms is Q-learning \cite{watkins1992q}, where the agent maintains an estimate of the Q-value function $\hat{Q}(s,a)$. When transitioning  from state-action pair $(s,a)$  to next state $s'$, the agent receives a payoff $r$ and updates the Q-function according to: 
\#\label{equ:Q_learning}
\hat{Q}(s,a)\leftarrow (1-\alpha)\hat{Q}(s,a)+\alpha\big[r+\gamma\max_{a'}\hat{Q}(s',a')\big],
\#
where  $\alpha>0$ is the stepsize/learning rate. Under certain conditions on $\alpha$, Q-learning can be proved to converge to the optimal Q-value function almost surely \cite{watkins1992q,szepesvari1999unified}, with  finite state and action spaces. Moreover, when combined with neural networks for function approximation, deep Q-learning has achieved great empirical breakthroughs in  human-level control  applications \cite{mnih2015human}. Another popular \emph{on-policy}  value-based method is SARSA, whose convergence  was established in {\cite{singh2000convergence}} for finite-space settings.  

An alternative while popular value-based RL algorithm is Monte-Carlo tree search (MCTS) \cite{chang2005adaptive,kocsis2006bandit,coulom2006efficient}, which estimates the optimal value function by constructing a search tree via Monte-Carlo simulations. Tree polices that 
judiciously select actions to balance exploration-exploitation are used to build and update the search tree. The most common tree policy is to apply the UCB1
(UCB stands for \emph{upper confidence bound})  algorithm, which was  originally devised for stochastic
multi-arm bandit problems \cite{agrawal1995sample,auer2002finite}, to each node of the tree. This yields the popular UCT algorithm \cite{kocsis2006bandit}.  Recent research endeavors on the non-asymptotic convergence of MCTS include \cite{pmlr-v80-jiang18a,shah2019reinforcement}. 

Besides, another significant task regarding value functions in RL is  to \emph{estimate the value function associated with a  given policy} (not only the optimal one). This task, usually referred to as \emph{policy evaluation}, has been tackled by  algorithms that follow a  similar update as \eqref{equ:Q_learning}, named  \emph{temporal difference} (TD) learning \cite{tesauro1995temporal,tsitsiklis1997analysis,sutton2018reinforcement}. Some other common policy evaluation algorithms with convergence guarantees include gradient TD methods  with linear \cite{sutton2008convergent,sutton2009fast,liu2015finite}, and nonlinear function approximations   \cite{bhatnagar2009convergent}. See \cite{dann2014policy} for a more detailed review on policy evaluation. 

\subsubsection{Policy-Based Methods} \label{sec:policyRL}
Another type of   RL algorithms directly searches over the policy space, which is usually estimated by parameterized function approximators like neural networks, namely, approximating  $\pi(\cdot\given s)\approx\pi_\theta(\cdot\given s)$.  As a consequence, the most  straightforward idea, which is to update the parameter along the gradient direction of the long-term reward, has been instantiated by the  policy gradient (PG) method.    
As a key premise for the idea, the closed-form of PG is given as \cite{sutton2000policy}  
\#\label{equ:policy_grad}
\nabla J(\theta)=\EE_{a\sim\pi_\theta(\cdot\given s),s\sim\eta_{\pi_\theta}(\cdot)}\big[Q_{\pi_\theta}(s,a)\nabla\log\pi_\theta(a\given s)\big],
\#
where $J(\theta)$ and $Q_{\pi_\theta}$ are  the expected return and Q-function under policy $\pi_\theta$, respectively,   $\nabla\log\pi_\theta(a\given s)$ is the score function of the policy, and $\eta_{\pi_\theta}$ is the state occupancy measure, either discounted or ergodic, under policy $\pi_\theta$.  
Then, various policy gradient methods, including  REINFORCE {\cite{williams1992simple}}, G(PO)MDP {\cite{baxter2001infinite}}, and actor-critic algorithms \cite{konda2000actor,bhatnagar2009natural}, have been proposed by estimating the gradient in different ways.  A similar idea also applies to deterministic policies in continuous-action settings, whose PG form has been derived recently  by \cite{silver2014deterministic}. 
Besides gradient-based ones, several other policy optimization methods have achieved state-of-the-art performance in many applications, including PPO  \cite{schulman2017proximal},  TRPO \cite{schulman2015trust}, soft actor-critic \cite{haarnoja2018soft}.

Compared with value-based RL methods, policy-based ones enjoy better convergence guarantees \cite{konda2000actor,yang2018finite,zhang2019global,agarwal2019optimality}, especially with neural networks for function approximation \cite{liu2019neural,wang2019neural}, which can readily handle massive or even continuous state-action spaces.  
Besides the value- and policy-based methods, there also exist RL algorithms based on the linear program formulation of an MDP; see recent efforts in \cite{chen2016stochastic,wang2017primal}.

\subsection{Multi-Agent RL Framework}\label{subsec:MARL_framework}

In a similar vein, multi-agent RL  also addresses  sequential decision-making problems, but with more than one agent involved. In particular, both the evolution of the system state and the reward received by each agent are influenced by the joint actions of all agents. More intriguingly, each agent has its own long-term reward to optimize, which now becomes a function of the  policies of all other agents. Such a general model finds broad applications in practice, see \S\ref{sec:applications} for a detailed review of several prominent examples.

\begin{figure*}[!t]
	\centering 
	\begin{tabular}{ccc} 
		\hskip-25pt\includegraphics[width=0.28\textwidth]{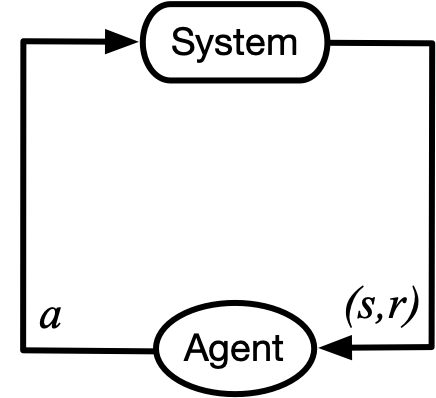}
		&
		\hskip9pt\includegraphics[width=0.31\textwidth]{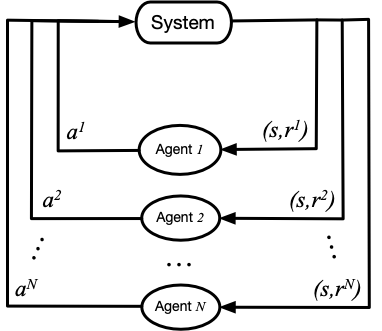}
		& 
		\hskip-17pt\includegraphics[width=0.44\textwidth]{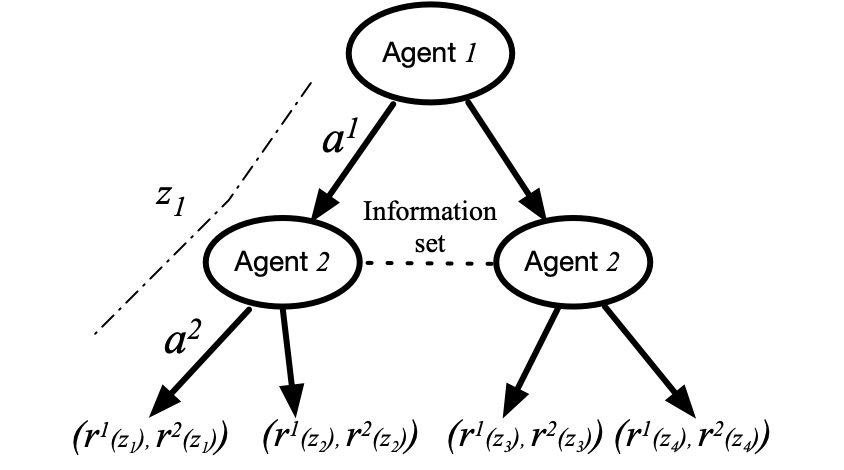}\\
		\hskip-15pt(a) Markov decision process  & \hskip -2pt(b) Markov game & \hskip-13pt (c) Extensive-form game	\end{tabular}
	\caption{Schematic diagrams for  the system evolution of a Markov decision process, a Markov game, and an extensive-form game, which correspond to the frameworks for single- and multi-agent RL, respectively. Specifically, in an MDP as in (a), the agent observes the state $s$ and receives reward $r$ from the system, after  outputting the action $a$; in an MG as in (b), all agents choose actions $a^i$ simultaneously, after observing the system state $s$ and receiving each individual reward $r^i$; in a two-player extensive-form game as in (c), the agents make decisions on choosing actions $a^i$ alternately, and receive each individual reward $r^i(z)$ at the end of the game, with $z$ being the terminal history. In the  imperfect information case, player $2$ is  uncertain about where he/she is in the game, which makes the information set non-singleton. } 
	\label{fig:model} 
\end{figure*}

In general, there exist two seemingly different but closely related theoretical frameworks for  MARL, Markov/stochastic games and extensive-form games, as to be introduced next. Evolution of the systems under different frameworks are illustrated in Figure \ref{fig:model}.


\subsubsection{Markov/Stochastic Games}\label{subsec:Markov_Games}
One direct generalization of MDP that captures the intertwinement of multiple agents is Markov games (MGs),   also known as stochastic games \cite{shapley1953stochastic}. 
Originated from the seminal work \cite{littman1994markov}, the framework of MGs\footnote{Similar to the single-agent setting, here we only introduce the infinite-horizon discounted setting for simplicity, though other settings of MGs, e.g., time-average-reward setting and finite-horizon episodic setting, also exist \cite{filar2012competitive}.} has long been used in the literature  to develop   MARL algorithms, see \S\ref{sec:algorithms} for more details.
We introduce the   formal definition as  below.

\begin{definition}\label{def:Markov_Game}
	A \emph{Markov game}  is defined by a tuple $(\cN,\cS,\{\cA^i\}_{i\in\cN},\cP,\{R^i\}_{i\in\cN},\gamma)$, where $\cN=\{1,\cdots,N\}$ denotes the set of  $N> 1$ agents, 
	$\cS$ denotes the state space observed by all agents, $\cA^i$ denotes the action space of agent $i$. Let $\cA:=\cA^1\times\cdots\times\cA^N$, then 
	 $\cP:\cS\times\cA\to\Delta(\cS)$ denotes the transition probability from any state $s\in\cS$ to any state $s'\in\cS$ for any joint action $a\in\cA$; $R^i:\cS\times\cA\times\cS\to\RR$ is the reward function that determines the immediate reward received by agent $i$ for a transition from $(s,a)$ to $s'$; $\gamma\in[0,1)$ is the discount factor. 
\end{definition}

At time $t$, each agent $i\in\cN$ executes an action $a^i_t$, according to the system state $s_t$. The system then transitions to state $s_{t+1}$, and rewards each agent $i$ by $R^i(s_t,a_t,s_{t+1})$. The goal of agent $i$ is to optimize its own long-term reward, by finding the policy $\pi^i:\cS\to\Delta(\cA^i)$ such that $a^i_t\sim \pi^i(\cdot\given s_t)$. 
As a consequence, the value-function $V^i:\cS\to\RR$ of agent $i$   becomes a function of the joint policy $\pi:\cS\to\Delta(\cA)$ defined as $\pi(a\given s):=\prod_{i\in\cN}\pi^i(a^i\given s)$. In particular, for any joint policy $\pi$ and state $s\in\cS$,  
\#\label{eq:V_pi}
V^i_{\pi^i,\pi^{-i}}(s):=\EE\bigg[\sum_{t\geq 0}\gamma^t R^i(s_t,a_t,s_{t+1})\bigggiven a^i_t\sim \pi^i(\cdot\given s_t),s_0=s\bigg],
\#
where $-i$ represents the indices of all agents in $\cN$ except agent $i$. 
Hence, the solution concept of an MG deviates  from that of an MDP, since the \emph{optimal} performance of each agent is controlled not only by its own  policy, but also the choices  of all other  players of the game. The most common solution concept, Nash equilibrium (NE)\footnote{Here, we focus only on stationary Markov Nash equilibria, for the infinite-horizon discounted MGs considered.}, is defined as follows \cite{basar1999dynamic,filar2012competitive}. 


\begin{definition}\label{def:NE}
	A \emph{Nash equilibrium} of the Markov game $(\cN,\cS,\{\cA^i\}_{i\in\cN},\cP,\{R^i\}_{i\in\cN},\gamma)$ is a joint policy $\pi^*=(\pi^{1,*},\cdots,\pi^{N,*})$, such that for any $s\in\cS$  and $i\in\cN$
	\$
	V^i_{\pi^{i,*},\pi^{-i,*}}(s)\geq V^i_{\pi^i,\pi^{-i,*}}(s),\quad\text{~~for any~~}\pi^i.
	\$
\end{definition}

Nash equilibrium characterizes an equilibrium  point $\pi^*$, from which none of the agents has any  incentive to deviate. In other words, for any agent $i\in\cN$, the policy $\pi^{i,*} $ is the \emph{best-response} of $\pi^{-i,*}$. As a standard learning goal for MARL, NE always exists for finite-space infinite-horizon discounted MGs \cite{filar2012competitive}, but 
may not be unique in general.  Most of the MARL algorithms are contrived to converge to such an equilibrium point, if it exists. 



The framework of Markov games is general enough to umbrella various MARL settings summarized below. 

\vspace{7pt}
\noindent {\bf Cooperative Setting:}
\vspace{5pt}

In a fully cooperative setting, all agents usually share a common reward function, i.e., $R^1=R^2=\cdots=R^N=R$. 
We note that this model is also  referred to as \emph{multi-agent MDPs} (MMDPs)  in the  AI community \cite{boutilier1996planning,lauer2000algorithm}, and \emph{Markov teams/team Markov games}   in the control/game theory community \cite{yoshikawa1978decomposition,ho1980team,wang2003reinforcement,mahajan2008sequential}. Moreover, from the game-theoretic perspective, this cooperative setting can also be viewed as a special case of Markov \emph{potential} games \cite{gonzalez2013discrete,zazo2016dynamic,valcarcel2018learning}, with the potential function being the common accumulated reward. 
With this model in mind, the value function and Q-function are identical to all agents, which thus enables the single-agent RL algorithms, e.g., Q-learning update \eqref{equ:Q_learning}, to be applied, if all agents are coordinated as one decision maker.  
 The global optimum for cooperation now constitutes a Nash equilibrium of the game.  

Besides the common-reward model, another slightly more general and surging model for cooperative MARL considers  \emph{team-average}  reward \cite{kar2013cal,zhang2018fully,doan2019finitea}. Specifically, agents are allowed to have different reward functions, which may be kept private to each agent, while the goal for cooperation is to optimize  the long-term reward corresponding to the average reward $\bar{R}(s,a,s'):=N^{-1}\cdot \sum_{i\in\cN}R^i(s,a,s')$ for any $(s,a,s')\in\cS\times\cA\times\cS$. The average-reward model, which allows more heterogeneity among agents, includes the model above as a special case.   It also   preserves   privacy among agents, and  facilitates  the development of  \emph{decentralized} MARL algorithms \cite{kar2013cal,zhang2018fully,wai2018multi}. Such heterogeneity also necessitates  the incorporation of \emph{communication} protocols  into MARL, and the analysis of communication-efficient MARL algorithms.

\vspace{7pt}
\noindent {\bf Competitive Setting:}
\vspace{5pt}

Fully competitive setting in MARL is typically modeled as \emph{zero-sum} Markov games, namely, $\sum_{i\in\cN}R^i(s,a,s')=0$ for any $(s,a,s')$. For ease of algorithm analysis and computational tractability, most literature focused on  
\emph{two} agents  that compete  against each other \cite{littman1994markov}, where clearly the reward of one agent is exactly the loss of the other. 
In addition to  direct applications to game-playing \cite{littman1994markov,silver2017mastering,OpenAI_dota_1v1}, zero-sum games also serve as a model for \emph{robust} learning, since the \emph{uncertainty} that impedes the learning process of the agent can be accounted for as 
a fictitious opponent in the game that is always against the agent   \cite{jacobson1973optimal,bacsar1995h,zhang2019policy}. Therefore, the Nash equilibrium yields a robust policy that optimizes the \emph{worst-case} long-term reward. 

\vspace{7pt}
\noindent {\bf Mixed Setting:}
\vspace{5pt}

Mixed setting is also known as the \emph{general-sum} game setting, where no  restriction  is imposed on the goal and relationship among agents \cite{hu2003nash,littman2001friend}. Each agent is self-interested, whose reward may be conflicting  with others'. Equilibrium solution  concepts from game theory, such as Nash equilibrium \cite{basar1999dynamic}, have  the most significant influence on algorithms that are developed for  this general setting.  Furthermore, we include the setting with both fully cooperative and competitive agents, for example, two zero-sum competitive teams with cooperative agents in each team \cite{lagoudakis2003learning,zhang2018finite,OpenAI_dota}, as instances of the mixed setting as well.


\subsubsection{Extensive-Form Games}

Even though they constitute  a classical formalism for MARL, Markov games can only handle the fully observed case, i.e., the agent has \emph{perfect information} on the system state $s_t$ and the executed action $a_t$ at time $t$. Nonetheless, a plethora of  MARL applications involve agents with only partial observability, i.e., \emph{imperfect information} of the game. 
Extension of Markov games to partially  observed case may be applicable, which, however, is challenging to solve, even under the cooperative setting \cite{oliehoek2016concise,bernstein2002complexity}.\footnote{Partially observed Markov games under the cooperative setting are usually formulated as decentralized POMDP (Dec-POMDP) problems. See \S\ref{subsec:partially_observed} for more discussions on this setting.}    
  
In contrast, another framework, named  \emph{extensive-form games} \cite{osborne1994course,shoham2008multiagent}, can handily model imperfect information for multi-agent decision-making. This framework is  rooted in   computational game theory  and has been  shown to admit polynomial-time algorithms under mild conditions \cite{koller1992complexity}. We briefly introduce the framework of extensive-form games as follows.    

\begin{definition} \label{def:ext_form_game}
	An \emph{extensive-form game}  is  defined by   $(\cN\cup \{c\},\cH,\cZ,\cA, \{ R^i\}_{i \in \cN},\tau,\pi^c,\cS)$, where $\cN = \{ 1, \ldots, N \}$ denotes the set of $N > 1 $ agents, and $c$ is a special agent called \emph{chance} or \emph{nature},  which has a fixed stochastic policy that specifies the randomness of the environment.  Besides,  $\cA$ is the set of all possible actions that agents can take and $\cH$ is the set of all possible \emph{histories}, where each history is  a sequence of actions taken  from the beginning of the game. 
	Let     $\cA(h)=\{a\given ha \in\cH\}$ denote  the set of  actions available after a nonterminal history $h $. 
	Suppose an agent takes action $a \in \cA(h)$ given  history $h \in \cH$, which then leads to a  new history $ha \in \cH$. 
	Among all histories, $\cZ\subseteq \cH$ is a subset of \emph{terminal histories} that represents the completion of a game. 
A utility is assigned to each agent $i \in \cN$ at a terminal history, dictated by the function 
	$R^i \colon  \cZ\rightarrow \RR$. 
	 Moreover, $\tau :\cH\to\cN\cup \{c\}$ is the \emph{identification} function that specifies which agent takes the action at each history. 
	If $\tau(h)=c$, the chance agent takes an  action $a$  according to its policy $\pi^c$, i.e., $a \sim \pi^c(\cdot\given h)$. Furthermore, $\cS$ is the partition of $\cH$ such that  
	 for any $s \in \cS$ and any $h, h' \in s $,  we have  $\tau(h) = \tau(h') $ and  $\cA (h) = \cA (h')$. In other words, histories $h$ and $h'$ in the same partition are indistinguishable to the agent that is about to take action, namely $\tau(h)$. The elements in $\cS$ are referred to as 
	 \emph{information states}.
	\end{definition}

Intuitively, the imperfect information of  
an extensive-form game is reflected by the fact that  agents cannot distinguish between histories in the same information set. 
Since we have $\tau(h) = \tau(h') $ and  $\cA (h) = \cA (h')$ for all $h,h'\in s$ and $s\in\cS$,  
for ease of presentation,  in the sequel, for all $h \in s$, 
we let $\cA(s) $ and $\tau(s)$ denote  $\cA (h)$ and $\tau(h)$, respectively. We also define a mapping $I \colon \cH \rightarrow \cS$ by letting $I (h) = s$ if $h \in s$. 
Moreover, we only consider games   where both  $\cH$  and $\cA$ are   finite sets. To simplify the notation, for any two histories $h, h' \in \cH$, we refer to $h$ as a \emph{prefix} of $h'$, denoted by $h  \sqsubseteq h'$,   if $h'$ can be reached from $h$ by taking a sequence of actions. In this case, we call $h'$ a \emph{suffix} of $h$. 
Furthermore, we assume throughout that 
the game features  \emph{perfect recall}, which  implies that  each agent remembers the sequence of the information states and  actions that have led to  its current information state.
The assumption of perfect recall is commonly made in the literature, which enables the existence of polynomial-time algorithms for solving the game \cite{koller1992complexity}. 
More importantly, by the celebrated 
Kuhn's theorem \cite{kuhn1953extensive}, under such an assumption, to find the set of  
Nash equilibria, it suffices to    restrict  the derivation  to the set of \emph{behavioral policies} which map each information set $s \in \cS$ to a probability distribution over $\cA (s)$. For any $i \in \cN$, let $\cS^i = \{ s\in \cS \colon \tau(s) = i \}$ be the set of information states of agent $i$. A joint policy of the agents is denoted by  $\pi = ( \pi^1, \ldots, \pi^N )  $, where  $\pi^i:\cS^i\to \Delta ( \cA (s) )$ is the policy of agent $i$. 
For any history $h$ and any joint policy $\pi$, we define the 
 \emph{reach probability} of $h$ under $\pi$  as 
\#\label{eq:reach_prod}
\eta _\pi (h) = \prod  _{h' \colon h'a\sqsubseteq h } \pi^{\tau(h')}    (a\given I(h')   ) =  \prod_{i \in \cN \cup \{c\}} \prod_{h' \colon h'a \sqsubseteq h, \tau(h') = i } \pi^i   (a \given I(h')  ),
\# 
 which specifies the probability that $h$ is created when all agents follow $\pi$. 
 We similarly define the reach probability of an information state $s $ under $\pi$ as $\eta_\pi (s) = \sum_{h \in s} \eta_\pi (h)$. The expected utility of agent $i \in \cN$ is  thus given by  $     \sum_{ z \in \cZ } \eta_\pi(z) \cdot R^i (z) $, which is denoted by $R^i ( \pi)  $ for simplicity. 
 Now we are ready to introduce  the solution concept for  extensive-form games, i.e., 
Nash equilibrium and its $\epsilon$-approximation, as follows. 

\begin{definition}\label{def:extens_NE}
	An \emph{$\epsilon$-Nash equilibrium} of an extensive-form game represented by $(\cN\cup \{c\},\cH,\cZ,\\\cA, \{ R^i\}_{i \in \cN},\tau,\pi^c,\cS)$  is a joint policy $\pi^*=(\pi^{1,*},\cdots,\pi^{N,*})$, such that for any 
	$i\in\cN$, 
	\$
	R^i (\pi^{i,*}, \pi^{-i,*})  \geq R^i ( \pi^i, \pi^{-i,*}) 
	 -\epsilon,\quad\text{~~for any policy~}\pi^i \text{~of agent }i.
	\$
	 Here   $\pi ^{-i}$ denotes the joint policy of    agents in  $\cN \setminus \{ i\} $ where agent $j$ adopts policy $\pi^j$ for all $j \in  \cN \setminus \{ i\}$.  Additionally, if $\epsilon=0$, $\pi^*$ constitutes a \emph{Nash equilibrium}. 
\end{definition}

\vspace{7pt}
\noindent {\bf Various Settings:}
\vspace{5pt}

Extensive-form games are in general used to model non-cooperative settings. 
Specifically, zero-sum/constant-sum utility with $\sum_{i\in\cN}R^i=k$ for some constant $k$ corresponds to the fully competitive setting; general-sum utility function results in the mixed setting.  
More importantly, settings of different information structures can also be characterized  by   extensive-form games. In particular, a \emph{perfect information} game is  one where each information set is a singleton, i.e., for any $s\in\cS$, $|s|=1$;  an \emph{imperfect information} game is  one where  there exists $s\in\cS$, $|s|>1$. In other words, with  imperfect information, the information state $s$ used for decision-making represents more  than one possible history, and the agent cannot distinguish between them. 

Among various settings, the zero-sum imperfect information setting has been the main focus of theoretical studies that bridge MARL and extensive-form games \cite{zinkevich2008regret,heinrich2015fictitious,srinivasan2018actor,omidshafiei2019neural}. It has also motivated MARL algorithms that  revolutionized competitive  setting applications like Poker AI \cite{rubin2011computer,brown2019superhuman}.

\vspace{7pt}
\noindent {\bf Connection to Markov Games:}
\vspace{5pt}

Note that the two formalisms in Definitions \ref{def:Markov_Game} and \ref{def:ext_form_game} are connected. In particular, for simultaneous-move Markov games, the choices of actions by other agents are unknown to an agent, which thus  leads  to different histories that can be aggregated as one information state $s$. Histories in these games are then  sequences of \emph{joint} actions, and the discounted accumulated reward instantiates the utility at the end of the game. 
Conversely, by simply setting $\cA^j=\emptyset$ at the state $s$ for agents $j\neq \tau(s)$, the extensive-form game reduces to a Markov game with \emph{state-dependent} action spaces. See \cite{lanctot2019openspiel} for a more detailed discussion on the  connection.


\begin{remark}[Other MARL Frameworks]
	{{Several other theoretical frameworks for MARL also exist in the literature, e.g.,  normal-form and/or  repeated games  \cite{claus1998dynamics,bowling2001rational,kapetanakis2002reinforcement,conitzer2007awesome}, and partially observed Markov games \cite{hansen2004dynamic,amato2013decentralized,amato2015scalable}. 
However, the former framework can be viewed as a special case of MGs, with a singleton state; most early theories of  MARL in this framework have been   restricted to small scale problems  \cite{bowling2001rational,conitzer2007awesome,kapetanakis2002reinforcement} only. 
MARL in the latter framework, on the other hand, is inherently challenging to address in general \cite{bernstein2002complexity,hansen2004dynamic}, leading to relatively scarce theories in the literature.   Due to space limitation,   we do not introduce these models here  in any detail. We will briefly review  MARL algorithms under some of these models, especially the partially observed setting, in    \S\ref{sec:algorithms}, though. Interested readers are referred to  the early review \cite{bu2008comprehensive} for more discussions on MARL in normal-form/repeated games. }}
\end{remark}

\section{Challenges in MARL Theory}\label{sec:challenges} 

Despite a general model that finds broad applications,  MARL  suffers  from several challenges in theoretical analysis, in addition to those that  arise in single-agent RL. We summarize below the  challenges that we regard as fundamental in developing theories for MARL.  
 

\subsection{Non-Unique  Learning Goals}\label{subsec:learning_goal}
Unlike single-agent RL, where the goal of the agent is to maximize the long-term return efficiently, the learning goals of MARL can be vague at times. In fact, as argued in \cite{shoham2003multi}, the \emph{unclarity} of the problems being addressed is the  fundamental flaw in many early MARL works. Indeed, the goals that need to be considered in the analysis of MARL algorithms can be multi-dimensional. The most common goal,  which has, however,  been challenged in \cite{shoham2003multi}, is the convergence to Nash equilibrium as defined in \S\ref{subsec:MARL_framework}. By definition, NE characterizes the point that no agent will deviate from, if any algorithm  \emph{finally} converges.  This is undoubtedly a  reasonable solution concept in game theory, under the assumption that the agents are all \emph{rational}, and are capable of perfectly reasoning and infinite mutual
modeling of agents. However, with \emph{bounded rationality}, the agents may only be able to perform \emph{finite} mutual modeling \cite{shoham2008multiagent}. As a result, the learning dynamics that are devised to converge to  NE may not be justifiable for practical MARL  agents. Instead, the goal may be focused on designing the best \emph{learning strategy} for a given agent and \emph{a fixed class of the other agents in the
game.}   In fact, these two goals are styled as \emph{equilibrium agenda} and \emph{AI agenda} in \cite{shoham2003multi}. 

Besides, it has  also been  controversial that \emph{convergence} (to the equilibrium point) is the dominant performance criterion for MARL algorithm analysis. In fact, it has been   recognized in \cite{zinkevich2006cyclic} that value-based MARL algorithms  fail to converge to the \emph{stationary} NE of general-sum Markov games, which motivated the new solution concept of \emph{cyclic equilibrium} therein, at which the agents  cycle rigidly through a set of stationary policies, i.e., not converging to any NE	policy. Alternatively, \cite{bowling2001rational,bowling2002multiagent}  separate the learning goal into being both \emph{stable} and \emph{rational}, where the former ensures the algorithm to be convergent, given  a predefined, targeted class of opponents' algorithms, while the latter requires the convergence to a best-response when the other agents remain stationary. If all agents are both stable and rational,  convergence to  NE naturally arises in this context.  
Moreover, the notion of \emph{regret}  introduces another angle to capture agents' rationality, which measures the performance of the algorithm compared to the best hindsight static strategy \cite{bowling2001rational,bowling2005convergence,blum2007learning}. No-regret algorithms with asymptotically zero average regret guarantee the convergence to the equilibria of certain games \cite{hart2001reinforcement,bowling2005convergence,zinkevich2008regret}, which in essence guarantee  that the agent is not \emph{exploited} by others. 

In addition to the goals concerning optimizing the return, several other goals that are special to multi-agent systems  
have also drawn increasing attention. For example, \cite{kasai2008learning,foerster2016learning,kim2019learning} investigate  \emph{learning to communicate}, in order for the agents to better coordinate. Such a concern on communication protocols has naturally inspired the recent studies on \emph{communication-efficient} MARL \cite{chen2018communication,lin2019communication,ren2019communication,kim2019message}. 
Other important goals  include how to learn without over-fitting certain agents \cite{he2016opponent,lowe2017multi,pmlr-v80-grover18a}, and  how to learn robustly  with either malicious/adversarial or failed/dysfunctional learning agents \cite{gao2018adversarial,li2019robust,zhang2019non}. 
Still in their infancy, some works concerning aforementioned goals provide  only empirical results, leaving plenty of room for theoretical studies.




\subsection{Non-Stationarity}\label{subsec:non_stationary}

Another key challenge of MARL lies in the fact that multiple agents usually learn concurrently, causing the  environment faced by each individual agent to be \emph{non-stationary}. 
In particular,  the action taken by one agent affects the reward of  other opponent agents, and the evolution of the state. 
As a result, the learning agent is required to account for how the other agents behave and adapt to the \emph{joint behavior} accordingly. 
This   invalidates the stationarity  assumption for establishing the convergence of single-agent RL algorithms, namely, the stationary Markovian property of the environment such that the individual reward and current state depend only on the previous state and action taken. This precludes the direct use of mathematical tools for single-agent RL analysis in MARL.

Indeed, theoretically, if the agent ignores this issue and optimizes its own policy assuming a stationary environment, which is usually referred to as an \emph{independent learner}, the algorithms may fail to converge \cite{tan1993multi,claus1998dynamics}, except for several special settings \cite{arslan2017decentralized,yongacoglu2019learning}. Empirically, however,  independent learning may achieve satisfiable performance in  practice  \cite{matignon2012independent,foerster2017stabilising}.  
As the most well-known issue in MARL, non-stationarity has long been recognized in 
the literature  \cite{bu2008comprehensive,tuyls2012multiagent}. A recent comprehensive survey \cite{hernandez2017survey} peculiarly 
provides an overview  on how  it is modeled and addressed by  state-of-the-art  multi-agent learning algorithms. We thus do not include any further discussion on this challenge, and refer interested readers to \cite{hernandez2017survey}. 

\subsection{Scalability Issue}\label{subsec:scala_issue}
To handle non-stationarity, each individual agent may need to account for the \emph{joint action space},  whose dimension  increases exponentially with the number of agents. This is also referred to as the \emph{combinatorial nature} of MARL \cite{hernandez2018multiagent}. Having a large  number of agents complicates the theoretical analysis, especially the convergence analysis, of MARL. This argument is   substantiated by the fact that  theories on  MARL for the two-player zero-sum setting are much more extensive and advanced than those for general-sum settings with more than two agents, see \S\ref{sec:algorithms} for a detailed comparison.  
One possible remedy for the scalability issue is to assume additionally the \emph{factorized}  structures of either the value or reward functions with regard to the action dependence; see  \cite{guestrin2002coordinated,guestrin2002multiagent,kok2004sparse} for the original heuristic ideas, and \cite{sunehag2018value,rashid2018qmix} for recent empirical progress. Relevant theoretical analysis had  not  been established until recently \cite{qu2019exploiting}, which considers a special dependence structure, and develops a provably convergent model-based  algorithm. 

Another  theoretical challenge of MARL that is brought about independently  of, but worsened by, the scalability issue,   is to build up theories for deep multi-agent RL. 
Particularly, scalability issues 
 necessitate the use of function approximation, especially  deep  neural networks,   in MARL. Though  empirically successful,  the theoretical analysis of deep MARL is an almost uncharted territory, with the currently limited understanding of deep 
learning theory, not alone the deep RL theory.  This is included as one of the future research directions in \S\ref{sec:conclusion}. 



\subsection{Various Information Structures}\label{subsec:info_structure}

\begin{figure*}[!t] 
	\centering
	\begin{tabular}{ccc}
		\hskip-10pt\includegraphics[width=0.28\textwidth]{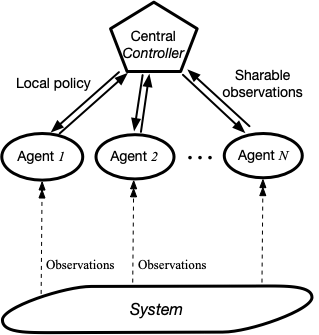}
		& 
		\hskip10pt\includegraphics[width=0.31\textwidth]{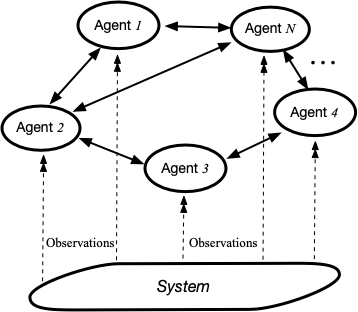}
		&  
		  \hskip10pt\includegraphics[width=0.31\textwidth]{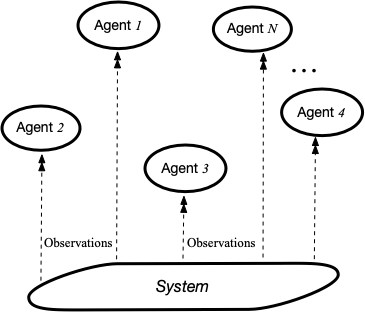}\\
		\hskip-15pt(a) Centralized setting   & \hskip 15pt \parbox{5cm}{(b) Decentralized setting \\ ~~~~~with networked agents} & \hskip10pt (c) Fully decentralized setting	\end{tabular}
	\caption{Three representative information structures in MARL. Specifically, in (a), there exists a central controller that can aggregate information from the agents, e.g., joint actions, joint rewards, and joint observations,  and even design policies for all agents. The information exchanged between the central controller and the agents can thus include both some private observations from the agents, and the local policies designed for each agent from the controller. In both (b) and (c), there is no such a central controller, and are thus both referred to as decentralized structures. In (b), agents are connected via a possibly time-varying communication network, so that the local information can spread across the network, by information exchange with only each agent's neighbors. (b) is more common in cooperative MARL settings. 
	In (c), the agents are full decentralized, with no explicit information exchange with each other. Instead, each agent makes  decisions based on its local observations, without any coordination and/or aggregation of data. The local observations that vary across agents,  however, may contain some global information, e.g., the joint actions of other agents, namely the control sharing information structure \cite{mahajan2013optimal}. Such a fully decentralized structure can also be found in many game-theoretic learning algorithms. } 
	\label{fig:info_struc} 
\end{figure*}

Compared to the single-agent case, the information structure of MARL, namely, \emph{who knows what} at the training and execution, is more involved. For example,  in the framework of Markov games, it suffices to observe the instantaneous state $s_t$, in order for each agent to make decisions, since the local policy $\pi^i$ mapping from $\cS$ to $\Delta(\cA^i)$ contains the equilibrium policy. On the other hand, for extensive-form games, each agent may need to recall the history of  past decisions, under the common perfect recall assumption. 
Furthermore, as self-interested agents, each agent can scarcely  access either the \emph{policy} or the rewards of the opponents, but at most the  action samples taken by them. This partial information aggravates the issues caused by non-stationarity, as the samples can hardly recover the exact behavior of the opponents' underlying policies, which increases the non-stationarity viewed by individual agents. The extreme case is the aforementioned \emph{independent learning} scheme, which assumes the observability of  only the local action and reward, and suffers from non-convergence in general \cite{tan1993multi}.

Learning schemes resulting  from various 
information structures lead to various levels of difficulty  for theoretical analysis. Specifically, to mitigate the partial information issue above, a great deal of work  assumes the existence of a \emph{central controller} that can 
collect information such as joint actions, joint rewards, and joint observations,  and even 
design policies for all agents \cite{hansen2004dynamic,oliehoek2014dec,lowe2017multi,foerster2017stabilising,gupta2017cooperative,foerster2018counterfactual,dibangoye2018learning,chen2018communication,rashid2018qmix}. This structure gives birth to the popular learning scheme of \emph{centralized-learning-decentralized-execution}, which stemmed  from the works on planning for the partially observed setting, namely, Dec-POMDPs  \cite{hansen2004dynamic,oliehoek2014dec,kraemer2016multi}, and has been widely adopted in recent (deep) MARL works \cite{lowe2017multi,foerster2017stabilising,gupta2017cooperative,foerster2018counterfactual,chen2018communication,rashid2018qmix}. For cooperative settings, this learning scheme greatly simplifies the analysis, 
allowing the use of tools for   single-agent RL analysis. Though, for non-cooperative settings with heterogeneous agents, this scheme does not significantly simplify the analysis, as the learning goals of the agents are not aligned, see \S\ref{subsec:learning_goal}.  

Nonetheless, generally such a central controller does not exist in many applications, except the ones that can easily access a  simulator, such as  video games and robotics. As a consequence, a  \emph{fully decentralized} learning scheme is preferred, which includes the aforementioned independent learning scheme as a special case. To address the non-convergence issue in  independent learning, agents are usually allowed to exchange/share local information with their neighbors over a  communication network  \cite{kar2013cal,macua2015distributed,macua2017diff,zhang2018fully,zhang18cdc,zhang2018finite,lee2018primal,wai2018multi,doan2019finitea,doan2019finiteb,suttle2019multi,lin2019communication}. We refer to this setting as \emph{a decentralized one  with networked agents}. 
Theoretical analysis for convergence is then made possible in this setting, the difficulty of which sits between that of single-agent RL and general MARL algorithms. 
Three different information structures are depicted  in Figure \ref{fig:info_struc}.


%
%
%
%

  
\section{MARL Algorithms with Theory}\label{sec:algorithms}
This section  provides a selective  review of MARL algorithms, and categorizes them according to the tasks to address.  Exclusively, we  review here the works that are focused on the theoretical studies only, which are mostly built upon the two representative MARL frameworks, fully observed Markov games and extensive-form games, introduced in \S\ref{subsec:MARL_framework}. A brief summary  on MARL for partially observed Markov games in  \emph{cooperative}  settings, namely, the Dec-POMDP problems, is also provided below in \S\ref{subsec:cooperative_MARL_alg},  due to their relatively  more mature   theory than that of MARL for general partially observed Markov games.
 
\subsection{Cooperative Setting}\label{subsec:cooperative_MARL_alg}

Cooperative MARL constitutes a great portion of  MARL settings, where all agents collaborate with each other to achieve some shared goal. Most cooperative MARL algorithms backed by  theoretical analysis are devised for the following more specific settings. 


\subsubsection{Homogeneous Agents}\label{subsubsec:homo_agents}

A majority of cooperative MARL settings  involve \emph{homogeneous} agents with a \emph{common} reward function that aligns all agents' interests. In the extreme case with large populations of agents, such a homogeneity also indicates that the agents play an \emph{interchangeable}  role in the system evolution, and can hardly be distinguished from each other. We elaborate more on  homogeneity below. 

\vspace{6pt}
\noindent{\bf Multi-Agent MDP \& Markov Teams}
\vspace{3pt}

Consider a Markov game as in Definition \ref{def:Markov_Game} with $R^1=R^2=\cdots=R^N=R$, where  the reward $R:\cS\times\cA\times\cS\to\RR$ is influenced by the joint action in $\cA=\cA^1\times\cdots\times \cA^N$. 
As a result, the Q-function is identical for  all agents. 
Hence, a straightforward algorithm proceeds by performing the standard 
Q-learning update \eqref{equ:Q_learning}  at each agent, but taking the $\max$ over the joint action space $a'\in\cA$. Convergence to the optimal/equilibrium Q-function has been established  in \cite{szepesvari1999unified,littman2001value}, when both state and action spaces are finite.

However,  convergence of the Q-function does not necessarily imply that of the equilibrium policy  for the Markov team, as any combination of  equilibrium policies extracted at each agent may not constitute an equilibrium policy, if the equilibrium policies are non-unique, and the agents fail to agree on which one to select. Hence, convergence to the NE policy is only guaranteed if either the equilibrium is assumed to be unique \cite{littman2001value}, or the agents are coordinated for equilibrium selection.  The latter idea has first been validated in the cooperative repeated games setting \cite{claus1998dynamics},  a special case of Markov teams with a singleton state, where  the agents are joint-action learners (JAL),  maintaining a Q-value for joint actions, and learning  empirical models of all others. Convergence to equilibrium point is claimed  in \cite{claus1998dynamics}, without a formal proof. 
For the actual Markov teams, this coordination has been exploited in  \cite{wang2003reinforcement}, which proposes \emph{optimal adaptive learning} (OAL), the first MARL algorithm   with provable  convergence to the equilibrium policy.  Specifically, OAL first learns the game structure, and constructs virtual games at each state that are \emph{weakly acyclic} with respect  to (w.r.t.) a biased set. OAL can be shown to converge to the NE, by introducing the biased adaptive play learning algorithm for the constructed  weakly acyclic games, motivated from \cite{young1993evolution}.

Apart from equilibrium selection, another subtlety   special to Markov teams (compared to single-agent RL) is the necessity to address the scalability issue, see  \S\ref{subsec:scala_issue}. As   independent Q-learning may fail to converge \cite{tan1993multi}, one early attempt toward developing  scalable while convergent algorithms for MMDPs is \cite{lauer2000algorithm}, which advocates a distributed Q-learning algorithm  that converges for \emph{deterministic} finite  MMDPs. Each agent maintains only a Q-table of state $s$ and local action $a^i$, and successively takes maximization over the joint action $a'$. No other agent's actions and their histories can be acquired by each individual agent. 
Several other heuristics (with no theoretical backing) regarding either reward or value function factorization  have been proposed to mitigate the scalability issue \cite{guestrin2002coordinated,guestrin2002multiagent,kok2004sparse,sunehag2018value,rashid2018qmix}.  Very recently,  \cite{pmlr-v97-son19a} provides a rigorous characterization of  conditions that justify  this value factorization idea. Another recent theoretical work  along this direction is \cite{qu2019exploiting},    which imposes a special dependence structure, i.e., a one-directional tree, so that the (near-)optimal policy of the overall MMDP can be provably well-approximated  by \emph{local policies}. 
More recently, 
\cite{yongacoglu2019learning}  has studied    common interest games, which  includes Markov teams as an example, and develops a \emph{decentralized} RL algorithm that relies on only states,  local   actions and rewards. With the same information structure as independent Q-learning \cite{tan1993multi}, the  algorithm is guaranteed to converge to \emph{team optimal} equilibrium policies, and not just equilibrium policies. This is important as in general, a suboptimal equilibrium can perform arbitrarily worse than
the optimal equilibrium  \cite{yongacoglu2019learning}. 


For 
policy-based methods,  
to  our knowledge, the only convergence guarantee  for this setting exists in  \cite{perolat2018actor}.  The authors propose two-timescale actor-critic \emph{fictitious play} algorithms, where at the slower timescale, the actor mixes the current policy and the best-response one w.r.t. the local Q-value estimate, while at the faster timescale the critic performs policy evaluation, as if all agents' policies are stationary. Convergence is established for  \emph{simultaneous move
multistage games} with a common (also zero-sum, see \S\ref{sec:policy_comp}) reward, a special Markov team  with initial and absorbing states, and each state being visited only once. 
   

\vspace{6pt}
\noindent{\bf Markov Potential Games}
\vspace{3pt}

From a game-theoretic perspective, a more general framework to embrace cooperation is \emph{potential games} \cite{monderer1996potential}, where there exits some \emph{potential} function shared by all agents, such that   if any agent changes its policy unilaterally, the change in its reward equals (or proportions to) that in the potential function. Though most potential games are stateless,  an extension named \emph{Markov potential games} (MPGs)  has gained increasing attention for modeling \emph{sequential} decision-making \cite{gonzalez2013discrete,zazo2016dynamic}, which includes Markovian states whose evolution is affected   by the joint actions. 
Indeed, MMDPs/Markov teams constitute   a particular case of MPGs, with the potential function being the common reward; such dynamic games can also be viewed as being {\em strategically equivalent} to Markov teams, using the terminology in, e.g., \cite[Chapter $1$]{basar2018handbook}. Under this model, \cite{valcarcel2018learning} provides  verifiable  conditions for a Markov game to be an MPG, and shows the equivalence between  finding closed-loop  NE in MPG and solving a single-agent optimal control problem. Hence,  single-agent RL algorithms are then enabled to solve this MARL problem.

\vspace{6pt}
\noindent{\bf Mean-Field Regime}
\vspace{3pt}

Another  idea toward tackling the scalability issue  is to take the setting to the \emph{mean-field} regime, with an extremely large number of homogeneous agents. 
Each agent's effect on the overall multi-agent system can thus become  infinitesimal, resulting in all agents being  interchangeable/indistinguishable. The interaction with other agents,  however, is captured simply by some mean-field quantity, e.g., the average state, or the empirical distribution of states. Each agent only needs to find the best response to the mean-field, which considerably simplifies the analysis.  
This mean-field view of multi-agent systems has been approached by the mean-field games (MFGs) model \cite{huang2003individual,huang2006large,lasry2007mean,bensoussan2013mean,tembine2013risk}, the team model with mean-field sharing \cite{arabneydi2014team,arabneydi2017new}, and the game model with mean-field actions \cite{pmlr-v80-yang18d}.\footnote{The difference between mean-field teams and mean-field games is mainly the solution concept: optimum versus equilibrium, as the difference between general dynamic team theory \cite{witsenhausen1971separation,yoshikawa1978decomposition,yuksel2013stochastic} and game theory \cite{shapley1953stochastic,filar2012competitive}. Although the former can be viewed as a special case of the latter, related works are usually reviewed separately in the literature.  We  follow here this convention. 
} 

MARL in these models have not been explored until recently, mostly in the non-cooperative setting of MFGs, see \S\ref{subsec:mixed_setting} for a more detailed review. Regarding the cooperative setting, recent work  \cite{subramanian2018reinforcement} studies RL for Markov teams with mean-field sharing \cite{arabneydi2014team,arabneydi2016linear,arabneydi2017new}. Compared to MFG, the model considers agents that share a common reward function depending only on the local state and the mean-field, which encourages cooperation among the  agents. Also, the term mean-field refers to the \emph{empirical average} for the states of \emph{finite} population, in contrast to the \emph{expectation} and \emph{probability distribution} of \emph{infinite} population in MFGs. 
 Based on the dynamic programming decomposition for the specified model  \cite{arabneydi2014team}, several popular  RL algorithms are easily translated to address this setting \cite{subramanian2018reinforcement}. More recently, \cite{carmona2019linear,carmona2019model}  approach the problem from a mean-field control (MFC) model,  to model large-population  of cooperative decision-makers. Policy gradient methods are proved to converge for linear quadratic MFCs in \cite{carmona2019linear}, and mean-field Q-learning is then shown to converge for general MFCs \cite{carmona2019model}.
 

\subsubsection{Decentralized Paradigm with Networked Agents}\label{subsubsec:networked_paradigm}
 
Cooperative agents in  numerous  
practical multi-agent systems are not always  homogeneous. Agents may have different preferences, i.e., reward functions, while  they still form a team to maximize the return of the \emph{team-average} reward $\bar R$, with $\bar R(s,a,s')=N^{-1}\cdot\sum_{i\in\cN}R^i(s,a,s')$. More subtly, the reward function is sometimes not sharable with others, as the preference is kept private to each agent.  
This setting finds broad applications in  engineering systems as sensor networks  \cite{rabbat2004distributed},  smart grid \cite{dall2013distributed,zhang2018dynamicpower}, intelligent transportation systems \cite{adler2002cooperative,zhang2018dynamic}, and robotics \cite{corke2005networked}.

Covering the homogeneous  setting in \S\ref{subsubsec:homo_agents} as a special case, the specified one definitely  requires more coordination, as, for example, the  global value function  cannot be estimated locally without knowing other agents' reward functions. With a central controller, most MARL algorithms reviewed in   \S\ref{subsubsec:homo_agents}  directly apply, since the controller can collect and average the rewards, and distributes the information to all agents. Nonetheless,  such a controller may not exist in 
most aforementioned applications,  due to  either cost, scalability, or robustness concerns \cite{rabbat2004distributed,dall2013distributed,zhang2018distributedauto}. 
Instead,   the agents may be able to share/exchange information with their neighbors over  a possibly time-varying and sparse communication network, as  illustrated in    Figure \ref{fig:info_struc} (b). 
Though MARL under this \emph{decentralized/distributed}\footnote{Note that hereafter  we use \emph{decentralized} and \emph{distributed} interchangeably for describing this paradigm. } paradigm  is   imperative, it is  relatively less-investigated, in comparison to the extensive   results on distributed/consensus algorithms  that solve \emph{static/one-stage} optimization problems \cite{nedic2009distributed,agarwal2011distributed,jakovetic2011cooperative,tu2012diffusion}, which, unlike RL, involves no system \emph{dynamics}, and does not maximize the \emph{long-term}  objective as a \emph{sequential-decision making} problem. 

\vspace{6pt}
\noindent{\bf Learning Optimal Policy}
\vspace{3pt}

The most significant goal is to learn the optimal joint policy, while each agent only accesses to  local and neighboring information over the network.  
The idea of MARL with networked agents dates back to \cite{varshavskaya2009efficient}. 
To our knowledge, 
 the first provably convergent MARL algorithm under this setting is  due to \cite{kar2013cal}, which incorporates  the idea of  \emph{consensus $+$ innovation} to the standard Q-learning algorithm, yielding the \emph{$\mathcal{QD}$-learning} algorithm with the following update
\$
Q^i_{t+1}(s,a)\leftarrow & Q^i_{t}(s,a)+\alpha_{t,s,a}\Big[R^i(s,a)+\gamma\max_{a'\in\cA}Q^i_t(s',a')-Q^i_t(s,a)\Big] \notag \\
& \qquad -\beta_{t,s,a}\sum_{j\in\cN^i_t}\big[Q^i_t(s,a)-Q^j_t(s,a)\big],
\$
where $\alpha_{t,s,a},~\beta_{t,s,a}>0$ are  stepsizes, $\cN^i_t$ denotes the set of neighboring agents of agent $i$, at time $t$.  Compared to the Q-learning update \eqref{equ:Q_learning}, \emph{$\mathcal{QD}$}-learning appends an innovation term that captures the difference of Q-value estimates from its neighbors. With certain conditions on the stepsizes, the algorithm is guaranteed to converge to the optimum Q-function for the tabular setting.  

Due to the scalability issue, function approximation is vital in MARL, which necessitates  the development of policy-based  algorithms. Our work  \cite{zhang2018fully} proposes actor-critic algorithms for this setting. Particularly, each agent parameterizes its own policy $\pi^i_{\theta^i}:\cS\to\Delta(\cA^i)$ by some parameter $\theta^i\in\RR^{m^i}$, the policy gradient of the return  is first derived as 
\#\label{eq:policy_gradient_thm}
		\nabla _{\theta^{i} } J({\theta}) &= \EE \left [ \nabla _{\theta^i} \log \pi _{\theta^i} ^i(s,a^i)\cdot  Q_{{\theta}}(s,a) \right ]
\#
\normalsize
where  $Q_{{\theta}}$ is the global value function corresponding to $\bar R$ under the joint policy $\pi_\theta$ that is defined as $\pi_\theta(a\given s):=\prod_{i\in\cN}\pi^i_{\theta^i}(a^i\given s)$. As an analogy to \eqref{equ:policy_grad}, the policy gradient in \eqref{eq:policy_gradient_thm}  involves the expectation of the product between  the local score function $\nabla _{\theta^i} \log \pi _{\theta^i} ^i(s,a^i)$ and the global  Q-function $Q_{{\theta}}$.  The latter, nonetheless, cannot be estimated individually at each agent. As a result, by parameterizing each local copy of $Q_{{\theta}}(\cdot,\cdot)$ as $Q_{{\theta}}(\cdot,\cdot;\omega^i)$ for agent $i$, we propose the following consensus-based TD learning for the critic step, i.e., for estimating $Q_{{\theta}}(\cdot,\cdot)$:
\#
\tilde{\omega}^i_{t}=\omega^i_{t}+\beta_{\omega,t}\cdot \delta^i_{t}\cdot \nabla_{\omega} Q_t(\omega^i_t),\qquad\qquad{\omega}^i_{t+1} = \sum_{j\in\cN}c_t(i,j)\cdot \tilde{\omega}^j_{t}
\label{equ:MARL_critic_1},
\#
where $\beta_{\omega,t}>0$ is the stepsize,  
$\delta^i_{t}$ is the local TD-error calculated using $Q_{{\theta}}(\cdot,\cdot;\omega^i)$. The first relation in \eqref{equ:MARL_critic_1} performs the  standard TD update, followed by a weighted  combination of the neighbors' estimates $\tilde{\omega}^j_{t}$. The weights here, $c_t(i,j)$, are  dictated by the topology of the communication network,  with non-zero values only if  two agents $i$ and $j$ are connected at time $t$. They also need to satisfy the \emph{doubly stochastic} property in expectation, so that $\omega^i_{t}$ reaches a \emph{consensual} value for all $i\in\cN$ if it converges. 
Then, each agent $i$ updates its policy following stochastic policy gradient given by \eqref{eq:policy_gradient_thm} in the actor step, using its own Q-function estimate $Q_{{\theta}}(\cdot,\cdot;\omega^i_t)$.  
A variant algorithm  is also introduced in  \cite{zhang2018fully}, relying on not the Q-function, but the state-value function approximation, to estimate the global advantage function. 

With these in mind, almost sure convergence  is established    in \cite{zhang2018fully} for these decentralized actor-critic algorithms, when     linear functions are  used for value function approximation.  
Similar ideas are also extended to the setting with continuous spaces \cite{zhang18cdc}, where deterministic policy gradient (DPG) method  is usually used. Off-policy exploration, namely a stochastic behavior policy, is required  for DPG, as the deterministic on-policy may not be explorative enough. However, in the multi-agent setting, as the  policies of other agents are unknown, the common off-policy approach for DPG \cite[\S 4.2]{silver2014deterministic} does not apply. Inspired by the expected policy gradient (EPG) method \cite{ciosek2018expected} which unifies stochastic PG (SPG) and DPG, we develop an algorithm that remains on-policy, but reduces the variance of general SPG  \cite{zhang18cdc}. In particular, we derive the multi-agent version of EPG, based on which we  develop  the actor step that can be implemented in a decentralized fashion, while the critic step still follows  \eqref{equ:MARL_critic_1}. Convergence of the algorithm is then also established when linear function approximation is used \cite{zhang18cdc}. In the same vein, \cite{suttle2019multi} considers the extension of \cite{zhang2018fully} to an off-policy  setting, building upon the  emphatic temporal differences (ETD) method for the critic \cite{sutton2016emphatic}. By incorporating the analysis of ETD($\lambda$) \cite{yu2015convergence} into \cite{zhang2018fully}, almost sure convergence guarantee has also been established. Another off-policy algorithm for the same setting is proposed concurrently by  \cite{zhang2019distributed}, where agents do not share their estimates of value function. Instead, the agents aim to reach consensus over the global optimal policy estimation. Provable convergence is then established for the algorithm, with  a  local critic and a consensus actor.

RL for decentralized networked agents has also been investigated in  \emph{multi-task}, in addition to the multi-agent, settings.  In some sense, the former can be regarded as a simplified version of the latter, where each agent deals with an \emph{independent MDP} that is not affected by other agents, while the goal is still to learn the  optimal joint policy that accounts for the average reward of all agents. 
\cite{pennesi2010distributed} proposes a distributed actor-critic algorithm,  assuming that the states, actions, and rewards are all local to each agent. Each agent performs a local TD-based critic step, followed by a consensus-based actor step that follows the gradient calculated using  information exchanged from the neighbors. Gradient of the average return is then proved to converge to zero as the iteration goes to infinity. \cite{macua2017diff} has  developed  \emph{Diff-DAC}, another distributed actor-critic algorithm for this setting, from duality theory. The updates resemble those  in \cite{zhang2018fully}, but provide additional insights that   actor-critic is  actually an instance of
the dual ascent method for solving  a linear program. 

Note that all the aforementioned convergence guarantees are \emph{asymptotic}, i.e., the algorithms converge as the iteration  numbers go to infinity, and are restricted to the case with linear function approximations. This fails to quantify the performance when  finite iterations and/or samples are used, not to mention when nonlinear functions such as deep neural networks are utilized. As an initial step toward \emph{finite-sample analyses} in this setting with  more \emph{general} function approximation, we consider   in \cite{zhang2018finite} the \emph{batch} RL algorithms \cite{lange2012batch}, specifically, decentralized variants of the fitted-Q iteration (FQI) \cite{riedmiller2005neural,antos2008fitted}. Note that we focus on FQI since it  motivates the celebrated deep Q-learning algorithm \cite{mnih2015human} when  deep neural networks are used for function approximation. 
We study FQI variants for both the cooperative setting with networked agents, and the competitive setting with two teams of such networked agents (see \S\ref{sec:value_comp} for more details). In the former setting,   all agents cooperate to iteratively update the global Q-function estimate, by fitting nonlinear least squares with the target values as the responses. In particular, let $\cF$ be the  function class for Q-function approximation, $\{(s_j,\{a^i_j\}_{i\in\cN},s_{j}')\}_{j\in[n]}$ be the batch transitions dataset of size $n$  available to all agents, $\{r^i_j\}_{j\in[n]}$ be the \emph{local} reward samples private to each agent, and $y^i_j=r^i_j+\gamma\cdot \max_{a\in\cA}Q^i_t(s_{j}',a)$  be the local target value, where $Q^i_t$ is agent $i$'s Q-function estimate at iteration  $t$. Then,  all agents cooperate to find a common Q-function estimate by  solving
\#\label{equ:FQI_coop}
\min_{ f\in \cF}~~ \frac{1}{N}\sum_{i\in\cN}\frac{1}{2n} \sum_{j=1}^n \bigl [ y^i_j - f(s_j, a_j^1, \cdots,a_j^N) \bigr ]^2.  
\#
Since $y^i_j$ is only known to agent $i$, the problem in \eqref{equ:FQI_coop} aligns with the formulation of \emph{distributed/consensus optimization} in \cite{nedic2009distributed,agarwal2011distributed,jakovetic2011cooperative,tu2012diffusion,hong2017stochastic,nedic2017achieving}, whose global optimal  solution can be achieved by the algorithms therein, if  $\cF$ makes  $\sum_{j=1}^n  [ y^i_j - f(s_j, a_j^1, \cdots,a_j^N)  ]^2$ convex for each $i$. This is indeed the case if $\cF$ is a linear function class. Nevertheless, with only a finite iteration of  distributed optimization algorithms (common in practice), agents may not reach exact consensus, leading to an error of each agent's Q-function estimate away from the actual optimum of \eqref{equ:FQI_coop}. Such an error also exists when nonlinear function approximation is used.  Considering this error caused by decentralized computation, we follow the \emph{error propagation} analysis stemming  from single-agent batch RL \cite{munos2007performance,munos2008finite,antos2008fitted,antos2008learning,farahmand2010error}, to establish the finite-sample performance of the proposed algorithms, i.e., how the accuracy of the algorithms output   depends on the function class $\cF$, the number of samples within each iteration $n$, and the number of iterations $t$.


\vspace{6pt}
\noindent{\bf Policy Evaluation}
\vspace{3pt}

Aside from control, a series of algorithms  in this setting focuses  on the policy evaluation task only, namely, the critic step of the actor-critic algorithms. With the policy fixed,  this task embraces a neater formulation,  as the sampling  distribution becomes stationary, and the objective becomes convex under linear function approximation. This facilitates the  finite-time/sample analyses, in contrast to  most control algorithms with only  {asymptotic guarantees}.   Specifically, under joint policy $\pi$, suppose each agent parameterizes the value function by a linear function class $\{V_\omega(s):=\phi^\top(s)\omega:\omega\in\RR^d\}$, where $\phi(s)\in\RR^d$ is the  feature vector at $s\in\cS$, and $\omega\in\RR^d$ is the vector of parameters. For notational convenience, let $\Phi:=(\cdots;\phi^\top(s);\cdots)\in\RR^{|\cS|\times d}$,  $D=\diag[\{\eta_\pi(s)\}_{s\in\cS}]\in\RR^{|\cS|\times|\cS|}$ be a diagonal matrix constructed using the   state-occupancy measure $\eta_\pi$, $\bar R^\pi(s)=N^{-1}\cdot \sum_{i\in\cN}R^{i,\pi}(s)$, where $R^{i,\pi}(s)=\EE_{a\sim\pi(\cdot\given s),s'\sim P(\cdot\given s,a)}[R^i(s,a,s')]$, and $P^\pi\in\RR^{|\cS|\times|\cS|}$ with the $(s,s')$ element being $[P^\pi]_{s,s'}=\sum_{a\in\cA}\pi(a\given s)P(s'\given s,a)$. 
Then, the objective of all agents is to jointly minimize the mean square projected Bellman error (MSPBE) associated with the team-average reward, i.e., 
\#\label{equ:MSPBE_joint}
\min_{\omega}\quad\texttt{MSPBE}(\omega):=\big\|\Pi_{\Phi}\big(V_\omega-\gamma P^\pi V_\omega -\bar R^\pi\big)\big\|^2_D=\big\|A\omega-b\big\|^2_{C^{-1}},
\#
where $\Pi_{\Phi}:\RR^{|\cS|}\to \RR^{|\cS|}$ defined as $\Pi_{\Phi}:=\Phi(\Phi^\top D\Phi)^{-1}\Phi^\top D$ is the projection operator onto subspace $\{\Phi\omega:\omega\in\RR^d\}$, $A:=\EE\{\phi(s)[\phi(s)-\gamma \phi(s')]^\top\}$, $C:=\EE[\phi(s)\phi^\top(s)]$, and $b:=\EE[\bar R^\pi(s)\phi(s)]$. By replacing the expectation with samples and using the 
Fenchel duality, the finite-sum  version of \eqref{equ:MSPBE_joint} can be re-formulated as a distributed saddle-point problem 
\#\label{equ:MSPBE_joint_finite_sum}
\min_{\omega}\max_{\lambda^i,i\in\cN}\quad\frac{1}{Nn}\sum_{i\in\cN}\sum_{j=1}^n 2(\lambda^i)^\top A_j\omega -2(b^i_j)^\top \lambda^i-(\lambda^i)^\top C_j\lambda^i
\#
where $n$ is the data size, $A_j,C_j$ and $b^i_j$ are empirical estimates of $A,C$ and $b^i:=\EE[R^{i,\pi}(s)\phi(s)]$, respectively, using sample $j$.  
Note that \eqref{equ:MSPBE_joint_finite_sum} is convex in $\omega$ and concave in $\{\lambda^i\}_{i\in\cN}$. The use of   MSPBE as an objective is standard in multi-agent policy evaluation \cite{macua2015distributed,lee2018primal,wai2018multi,doan2019finitea,doan2019finiteb}, and the idea of  saddle-point reformulation  has been adopted in \cite{macua2015distributed,lee2018primal,wai2018multi,cassano2018multi}. Note that in \cite{cassano2018multi}, a variant of MSPBE, named H-truncated $\lambda$-weighted MSPBE, is advocated, in order to control the bias of the solution deviated from the actual mean square Bellman error minimizer.  

With the formulation \eqref{equ:MSPBE_joint} in mind, \cite{lee2018primal} develops a distributed variant of the gradient TD-based method, with asymptotic   convergence established using the ordinary differential equation (ODE) method.  In \cite{wai2018multi},  a double averaging scheme that combines the dynamic consensus  \cite{qu2017harnessing} and the SAG algorithm \cite{schmidt2017minimizing} has been proposed to solve the saddle-point problem \eqref{equ:MSPBE_joint_finite_sum} with a linear rate. \cite{cassano2018multi} incorporates the idea of   variance-reduction, specifically, AVRG in \cite{ying2018convergence}, into  gradient TD-based policy evaluation. Achieving the same linear rate as \cite{wai2018multi}, three  advantages are claimed in  \cite{cassano2018multi}: i)  data-independent memory requirement; ii) use of eligibility traces \cite{singh1996reinforcement}; iii) no need for  synchronization  in sampling. More recently, standard TD learning \cite{tesauro1995temporal}, instead of gradient-TD, has been generalized to this MARL setting, with special focuses on  finite-sample analyses \cite{doan2019finitea,doan2019finiteb}.  
Distributed TD($0$) is first studied in \cite{doan2019finitea}, using the proof techniques originated in \cite{pmlr-v75-bhandari18a}, which requires a projection on the iterates, and the data samples to be independent and identically distributed (i.i.d.). Furthermore, motivated by the recent progress in \cite{pmlr-v99-srikant19a}, finite-time performance of the  more general  distributed TD($\lambda$) algorithm is provided in \cite{doan2019finiteb}, with neither projection nor i.i.d. noise assumption needed.

Policy evaluation for networked agents has also been investigated under the setting of \emph{independent} agents interacting with \emph{independent} MDPs.   \cite{macua2015distributed} studies off-policy evaluation based on  the importance sampling technique. With no coupling among MDPs, an agent  does not need to know the actions of the other agents.  Diffusion-based distributed GTD is then proposed, and is shown to convergence in the mean-square sense with a sublinear rate. In  \cite{stankovic2016multi}, two variants of the TD-learning, namely, GTD2 and TDC \cite{sutton2009fast}, have been  designed for this setting, with weak convergence proved by the general stochastic approximation theory in \cite{stankovic2016distributed}, when agents are connected by a time-varying communication network. Note that \cite{cassano2018multi} also considers the independent MDP setting, with the same results established as the actual MARL setting.

%
%
%



\vspace{6pt}
\noindent{\bf Other Learning Goals}
\vspace{3pt}

Several other learning goals have also been explored for decentralized MARL with networked agents. \cite{zhang2016data} has considered  the \emph{optimal consensus}  problem, where each agent over the network tracks the states of its neighbors' as well as a leader's, so that the  consensus error is minimized by the joint policy. A policy iteration algorithm is then introduced,  followed by a practical  actor-critic algorithm using neural networks for function approximation. A similar consensus error objective is also adopted in \cite{zhang2018model}, under  the name of \emph{cooperative multi-agent graphical games}. A centralized-critic-decentralized-actor scheme is utilized for developing off-policy RL algorithms. 

Communication efficiency, as a key ingredient in the algorithm design for this setting, has drawn increasing attention recently \cite{chen2018communication,ren2019communication,lin2019communication}.  Specifically, \cite{chen2018communication} develops Lazily
Aggregated Policy Gradient (LAPG), a distributed PG algorithm that can reduce  the communication rounds between the agents and a central controller, by judiciously designing communication trigger rules. \cite{ren2019communication} addresses the same policy evaluation problem as \cite{wai2018multi}, and develops a   hierarchical distributed algorithm by proposing a mixing matrix different from the doubly stochastic one used in \cite{zhang2018fully,wai2018multi,lee2018primal}, which allows unidirectional information exchange among agents to save  communication. In contrast, the distributed actor-critic algorithm in  \cite{lin2019communication} reduces the communication by transmitting only   one scalar entry of its state vector at each iteration, while preserving provable  convergence as in \cite{zhang2018fully}.



\subsubsection{Partially Observed Model}\label{subsec:partially_observed}

We complete the overview for cooperative settings  by briefly introducing a class of significant but challenging models where agents are faced with partial  observability. Though common in practice, theoretical analysis of MARL algorithms in this setting is still relatively scarce, in contrast to the aforementioned fully observed settings.  
In general, this setting can be modeled by a decentralized POMDP (Dec-POMDP) \cite{oliehoek2016concise}, which shares almost all elements such as the reward function and the transition model, as the MMDP/Markov team model in \S\ref{subsec:Markov_Games}, except that each agent now only has its local observations of the system state $s$. With no accessibility  to other agents' observations, an  individual agent cannot maintain a global belief state, the sufficient statistic for decision making in single-agent POMDPs. Hence, Dec-POMDPs have been known to be  NEXP-complete \cite{bernstein2002complexity},    requiring super-exponential time to solve in the worst case.  

There is an increasing interest in developing planning/learning algorithms for Dec-POMDPs. 
Most of the algorithms are based on the \emph{centralized-learning-decentralized-execution} scheme. In particular, the  decentralized problem is first reformulated as a centralized one, which can be solved at a central controller with (a simulator that generates) the observation data of all agents. The policies are then optimized/learned using data, and distributed to all agents for  execution.   
Finite-state controllers (FSCs) are commonly used to represent  the local policy at each agent \cite{bernstein2009policy,amato2010optimizing}, which map local observation histories to actions. 
A Bayesian nonparametric approach is proposed in \cite{liu2015stick} to   determine the controller size of variable-size FSCs. 
To efficiently solve the centralized problem, a series of \emph{top-down}  algorithms have been proposed. In \cite{oliehoek2014dec},   the Dec-POMDP is converted  to  \emph{non-observable MDP} (NOMDP), 
a kind of centralized sequential decision-making problem, which is then  addressed by some  heuristic  tree search algorithms. 
As an extension of the NOMDP conversion, 
\cite{dibangoye2016optimally,dibangoye2018learning} convert Dec-POMDPs to   \emph{occupancy-state} MDPs (oMDPs), where the occupancy-states are distributions over hidden states and joint histories of observations. As the value functions of  oMDPs enjoy the piece-wise linearity and convexity, both tractable planning \cite{dibangoye2016optimally} and value-based learning \cite{dibangoye2018learning} algorithms have been developed. 
  
To further improve  computational  efficiency, 
several sampling-based planning/learning algorithms have also been proposed.  {In particular, } Monte-Carlo sampling {with} policy iteration and {the expectation-maximization} algorithm, are proposed in \cite{wu2010rollout} and \cite{wu2013monte},  respectively. 
Furthermore, Monte-Carlo tree search   has been applied {to} special classes of Dec-POMDPs, such as  multi-agent POMDPs \cite{amato2015scalable} and multi-robot active perception \cite{best2018dec}.  
In addition, policy gradient-based algorithms can also be developed for this centralized learning scheme \cite{dibangoye2018learning}, with a centralized critic and a decentralized actor.  
Finite-sample analysis can also be   established under this scheme \cite{amato2009achieving,banerjee2012sample}, for tabular settings with finite state-action spaces. 

Several attempts have also been  made to  enable \emph{decentralized learning} in Dec-POMDPs. When the agents share some common information/observations, \cite{nayyar2013decentralized} proposes to reformulate the problem as a centralized POMDP, with the common information being the observations of a \emph{virtual} central controller. This way, the centralized POMDP can be solved individually by each agent. In \cite{arabneydi2015reinforcement}, the reformulated POMDP has been  approximated by finite-state MDPs with exponentially decreasing approximation error, which are then solved by  Q-learning. Very  recently, \cite{zhang2019online} has developed a tree-search based algorithm for solving this centralized POMDP, which, interestingly, echoes back the heuristics for solving Dec-POMDPs directly as in \cite{amato2015scalable,best2018dec}, but with a more solid theoretical footing. Note that in both \cite{arabneydi2015reinforcement,zhang2019online}, a common random number generator  is used for all agents, in order to avoid communication among agents and enable a decentralized learning scheme.



\subsection{Competitive Setting}

 Competitive settings are usually modeled as \emph{zero-sum} games. Computationally, 
  there exists a great barrier between solving  two-player and multi-player zero-sum games. 
 In particular, even the  simplest   three-player   matrix games, are known to be 
 PPAD-complete \cite{papadimitriou1992inefficient,daskalakis2009complexity}.
 Thus, most existing results on competitive MARL  focus on two-player zero-sum games, with $\cN = \{1, 2\}$ and $R^1 + R^2 = 0$ in Definitions \ref{def:Markov_Game}  and \ref{def:ext_form_game}. 
 In the rest of this section, we review methods that provably find  a Nash (equivalently, saddle-point)  equilibrium in two-player Markov   or  extensive-form games.  
 The existing algorithms can mainly  be categorized into two classes: value-based  and policy-based approaches, 
 which are introduced separately in the sequel.

 \subsubsection{Value-Based Methods} \label{sec:value_comp}
 Similar as in single-agent MDPs, value-based methods aim to find an optimal value function  from which the joint  Nash equilibrium policy can be extracted.  
 Moreover, the optimal value function is known to be the unique fixed point of a Bellman operator, which can be obtained via dynamic programming type methods.
  
 Specifically, for simultaneous-move Markov games,  the value function defined in \eqref{eq:V_pi} satisfies  $V_{\pi^1, \pi^2}^1 = - V^2 _{\pi^1, \pi^2}   $. 
 and thus any   Nash equilibrium  $\pi^* = (\pi^{1,*}, \pi^{2,*} )$ satisfies 
 \#\label{eq:zero-sum-eq}
 V  _ {\pi^1, \pi^{1,*}} ^1 (s) \leq V  _ {\pi^{1,*}, \pi^{2,*}} ^1 (s)  \leq V^1   _ {\pi^{1,*}, \pi ^2} (s), \qquad \text{for any~}\pi = (\pi^1, \pi^2) ~\text{and}~s\in \cS. 
 \#
Similar as the  minimax theorem  in normal-form/matrix zero-sum games \cite{von2007theory}, for two-player zero-sum Markov games with finite state and action spaces,  one can define the  optimal value function $V^* \colon \cS \rightarrow \RR$  as
\#\label{eq:minimax}
V^* =   \max_{\pi^1} \min _{\pi^2} V^1 _{\pi^1, \pi^2}  = \min _{\pi^2} \max_{\pi^1}  V^1 _{\pi^1, \pi^2},
\#
Then \eqref{eq:zero-sum-eq} implies that 
$V_{\pi^{1,*}, \pi^{2,*}} ^1$  coincides with $V^*$ and any pair of policies $\pi^1$ and $\pi^2$ that attains  
 the supremum and infimum in \eqref{eq:minimax} constitutes a Nash equilibrium.
 Moreover, similar to MDPs, \cite{shapley1953stochastic} shows that $V^*$ is the unique solution of a Bellman equation and a Nash equilibrium can be constructed based on $V^*$. 
 Specifically,
 for any $V \colon \cS \rightarrow \RR$ and any $s \in \cS$, we define 
 \#\label{eq:Q_V}
 Q_V(s,a^1, a^2)  = \EE_{s' \sim \cP(\cdot \given s, a^1, a^2)} \big [R^1 (s, a^1, a^2, s' ) + \gamma \cdot V(s') \bigr ], 
 \#
where $Q_V(s,\cdot,\cdot)$ can be regarded as a matrix in $\RR^{| \cA^1|  \times | \cA^2|}$. Then we define the Bellman operator  $\cT^*$ by solving a  matrix zero-sum game regarding  $Q_V(s, \cdot, \cdot )$ as the payoff matrix, i.e., for any $s\in \cS$,  one can define 
 \#\label{eq:bellman_eq}
( \cT^* V) (s) = \texttt{Value} \bigl[ Q_V(s,\cdot , \cdot ) \bigr ]  = \max_{u \in \Delta(\cA^1) }\min _{v\in \Delta(\cA^2) }  \sum_{a\in \cA^1} \sum_{b\in \cA^2} u_{a} \cdot  v_b  \cdot Q _V(s,a, b), 
 \# 
 where we use $ \texttt{Value} (\cdot) $ to denote the optimal value of a matrix zero-sum game, which can be obtained by solving a  linear program  \cite{vanderbei2015linear}. 
 Thus,  the Bellman operator $\cT^*$ is $\gamma$-contractive in the $\ell_{\infty}$-norm and $V^*$ in \eqref{eq:minimax} is the unique solution to the Bellman equation $V = \cT^* V$. Moreover, letting $ p_1(V), p_2(V)$ be any solution to the 
  optimization problem in  \eqref{eq:bellman_eq}, we have that 
 $ \pi^* = (p_1(V^*), p_2(V^*) ) $ is a Nash equilibrium specified by Definition \ref{def:NE}. Thus, 
based on the Bellman operator $\cT^*$, \cite{shapley1953stochastic} proposes  the value iteration algorithm, which creates a sequence of value functions $\{ V_{ t }\}_{t \geq 1} $ satisfying $V_{t+1} = \cT ^* V_{t} $ that converges to $V^*$ with a linear rate. Specifically, we have 
 \$
 \| V_{t+1}  - V^* \|_{\infty} = \| \cT^* V_{ t  } -  \cT^* V^*  \|_{\infty} \leq \gamma \cdot  \|V_{t} -    V^* \|_{\infty} \leq \gamma ^{t+1} \cdot  \| V_{0} - V^* \|_{\infty}  . 
 \$
 In addition, a value iteration update  can be decomposed into the two steps.  In particular, 
  letting $\pi^1  $ be any   policy of player $1$ and  $V$ be any value function, we define Bellman operator $\cT^{\pi^1 }$ by 
 \#\label{eq:bellman_oper2}
 ( \cT^{\pi^1} V)(s) =  \min _{v\in \Delta(\cA^2) }  \sum_{a\in \cA^1} \sum_{b\in \cA^2} \pi^1(a\given s ) \cdot v_b\cdot  Q _V(s,a, b),
 \# 
 where $Q_V$ is defined in \eqref{eq:Q_V}. 
 Then we can equivalently  write a  value iteration update as 
\#\label{eq:value_iteration}
\mu_{t+1} = p_1(V_{t}) \qquad \text{and}\qquad   V_{ t+1 }= \cT^{\mu_{ t+1}} V_{t}. 
 \# 
 Such a decomposition 
 motivates the policy iteration  algorithm for two-player zero-sum games, which has been  studied in, e.g.,   \cite{hoffman1966nonterminating, van1978discounted, rao1973algorithms, patek1997stochastic, hansen2013strategy}, for different variants of such Markov games. 
 In particular,  from the perspective of player $1$, policy iteration creates a sequence $\{ \mu_{ t} , V_{ t }\}_{t\geq 0} $ satisfying  
 \#\label{eq:policy_iteration}
 \mu_{ t+1 } = p_1 ( V_{t  } ) \qquad \text{and} \qquad V_{ t+1 }  =(\cT^{\mu_{ t+1 }})^{\infty} V_{ t} ,
 \#
i.e.,    $V_{ t+1 } $ is the fixed point of   $\cT^{\mu_{ t+1 }}$. The updates for  player $2$ can be similarly constructed. By the definition in \eqref{eq:bellman_oper2}, the Bellman operator  $ \cT^{ \mu_{ t+1 }}$  is $\gamma$-contractive and its fixed point corresponds to  the value function associated with $(  \mu_{ t+1 }, \texttt{Br}  ( \mu_{ t+1 }) )$, where $ \texttt{Br}  ( \mu_{ t+1 }) $ is the best response policy of player $2$ when player $1$ adopts $ \mu_{ t+1 }$.  Hence, in each step of policy iteration, the player first finds an improved policy  $ \mu_{ t+1 }$ based on the current  function $V_{t}$, and then obtains a conservative value function by assuming that the opponent plays the best counter policy $ \texttt{Br} (\mu_{t+1}) $. It has been  shown in  \cite{hansen2013strategy} that the  value function sequence  $\{ V_{t} \}_{t\geq 0}$ monotonically increases to $V^*$  with a linear rate of convergence for turn-based zero-sum Markov games. 
 
 Notice  that both the  value and  policy iteration algorithms are model-based due to the need of computing the Bellman operator $\cT^{ \mu_{ t+1 }}$ in \eqref{eq:value_iteration} and \eqref{eq:policy_iteration}. 
By estimating the Bellman operator
via data-driven approximation,  
 \cite{littman1994markov} has proposed   minimax-Q learning, which extends the well-known  Q-learning algorithm  \cite{watkins1992q} for MDPs to zero-sum Markov games. In particular,  minimax-Q learning is an online, off-policy, and tabular method 
 which updates the         action-value function    $Q \colon \cS\times \cA \rightarrow \RR$ based on transition data $\{(s_t, a_t, r_t, s_t')\}_{t\geq 0}$, where $s_t' $ is the next state following $(s_t, a_t)$ and $r_t$ is the reward.   
 In the $t$-th iteration, it only updates  the value  of $Q(s_t, a_t) $    and keeps other entries of $Q$   unchanged. Specifically, we have 
   \#\label{eq:minimaxQ}
   Q  (s_t,a^1_t, a^2_t  ) \leftarrow   ( 1- \alpha_t )  \cdot Q (s_t,a^1_t, a^2_t   ) +\alpha_t  \cdot \bigl \{  r_t + \gamma \cdot \texttt{Value} \big[ Q  (s_t', \cdot,  \cdot) \bigr ]  \bigr \} ,
   \#
   where $\alpha_t \in (0,1)$ is the stepsize.
 {As shown in \cite{szepesvari1999unified}, under  conditions similar to those for   single-agent Q-learning \cite{watkins1992q}, function $Q$ generated by \eqref{eq:minimaxQ}   
   converges to the optimal action-value function $Q^* = Q_{V^*}$ defined by combining \eqref{eq:minimax} and \eqref{eq:Q_V}.} 
   Moreover, with a slight abuse of notation, if we define the  Bellman operator $\cT^*$ for action-value functions by  
    \#\label{eq:minimax_Q}
   ( \cT^* Q )(s,a^1, a^2 )  = \EE _{  s' \sim \cP(\cdot \given s, a^1, a^2)} \big  \{ R^1 (s, a^1, a^2, s' ) + \gamma \cdot {\tt{Value}}\bigl[  Q(s', \cdot, \cdot) \bigr ] \bigr \}, 
   \# 
  then we have  
   $Q^*$ 
  as the unique fixed point of $\cT^*$. 
  Since the target value 
  $ r_t + \gamma \cdot   \texttt{Value} [ Q  (s_t', \cdot,  \cdot) ] $ in \eqref{eq:minimaxQ} is an  estimator of $(\cT^* Q) (s_t, a_t^1, a_t^2)$, 
   minimax-Q learning can be viewed as a stochastic approximation algorithm for computing the fixed point of $\cT^*$. 
   Following \cite{littman1994markov}, 
  minimax-Q learning has been further extended to the function approximation setting  where $Q$ in \eqref{eq:minimaxQ} is approximated by a class of parametrized functions.  However, convergence guarantees for this minimax-Q learning with even linear function approximation have not been well understood. 
   Such a linear value function approximation also applies to a significant class  of zero-sum MG instances with continuous state-action spaces, i.e., linear quadratic (LQ) zero-sum games \cite{bacsar1995h,al2007adaptive,al2007model}, where  the reward function is quadratic with respect to the states and actions, while the transition model follows linear dynamics. In this setting, Q-learning based algorithm can be  guaranteed  to converge  to the NE \cite{al2007model}.

   To embrace general function classes, the framework of batch RL \cite{lagoudakis2002value,munos2007performance,munos2008finite,antos2008fitted,antos2008learning,farahmand2016regularized} can be adapted to  the multi-agent settings, as in the recent works  \cite{yang2019theoretical,zhang2018finite}.      As mentioned  in \S\ref{subsubsec:networked_paradigm} for cooperative batch MARL,   each agent  iteratively  updates the Q-function by fitting least-squares using  the target values.  Specifically, let $\cF$ be the function class of interest and let $  \{ (s_i, a_i^1, a_i^2, r_i, s_i' ) \}_{i\in[n]}$ be the dataset. For any $t \geq 0$,   let $Q_{t}$ be the current iterate in the $t$-th iteration, and  define $y_i = r_i + \gamma \cdot \texttt{Value} [ Q_{t} (s_i', \cdot , \cdot )]$ for all $i \in [n]$. Then we update $ Q_{t}$   by solving a least-squares regression problem in $\cF$, that is, 
  \#\label{eq:FQI}
  Q_{t+1} \leftarrow \argmin_{ f\in \cF} \frac{1}{2n} \sum_{i=1}^n \bigl [ y_i - f(s_i, a_i^1, a_i^2 ) \bigr ]^2.
  \#
  In such a two-player zero-sum Markov game setting, 
a finite-sample error bound on the Q-function estimate  is established in    \cite{yang2019theoretical}.

Regarding other finite-sample analyses, 
very recently, \cite{jia2019feature} has studied   zero-sum turn-based stochastic games (TBSG), a simplified zero-sum MG when the transition  model is embedded in some feature space and a generative model is available. Two Q-learning based  algorithms have been  proposed and analyzed  for this setting.  \cite{sidford2019solving} has proposed   algorithms that achieve near-optimal sample complexity for general zero-sum TBSGs with a generative model, by extending the previous near-optimal Q-learning algorithm for MDPs \cite{sidford2018near}.   
In the online setting, where the learner controls only one of the players that plays against an arbitrary opponent,  
\cite{wei2017online} has proposed UCSG, an algorithm for the  \emph{average-reward} zero-sum  MGs, using the principle of \textit{optimism in the  face of uncertainty} \cite{auer2007logarithmic, jaksch2010near}.  UCSG is shown to achieve a sublinear regret  compared to the game value when competing with an arbitrary opponent, and also achieve $\tilde O(\text{poly}(1/\epsilon))$ sample complexity if the opponent plays an optimistic best response.

 Furthermore, when it comes to zero-sum games  with imperfect information, \cite{kollery1994fast, von1996efficient, koller1996efficient, von2002computing} have proposed to transform  extensive-form games into normal-form games using the 
 \textit{sequence form} representation, which enables equilibrium  finding via linear programming. 
 In addition,   by lifting the state space to  the space of belief states,  \cite{parr1995approximating,rodriguez2000reinforcement,hauskrecht2000value, hansen2004dynamic,buter2012dynamic} have applied dynamic programming methods to zero-sum stochastic games. 
 Both of these approaches  guarantee  finding of a Nash equilibrium but are only efficient for small-scale problems. 
 Finally,  MCTS with UCB-type action selection rule  \cite{chang2005adaptive,kocsis2006bandit,coulom2006efficient} can also be applied to two-player turn-based games with incomplete  information \cite{kocsis2006bandit, cowling2012information, teraoka2014efficient,whitehouse2014monte,  kaufmann2017monte}, which lays the foundation for the recent success of deep RL for the game of Go \cite{silver2016mastering}. Moreover, these methods are 
 shown to converge to the minimax solution of the game, thus can be viewed as a counterpart of minimax-Q learning with Monte-Carlo sampling. 

  \subsubsection{Policy-Based Methods} \label{sec:policy_comp}
 
 Policy-based reinforcement learning methods introduced in \S\ref{sec:policyRL}
 can  also be extended to the multi-agent setting. Instead of finding the fixed point of the Bellman operator, a fair amount of methods only focus on a single agent and aim to  maximize the expected return  of that agent, disregarding  the other agents' policies.
 Specifically, 
 from the perspective of a single agent, the environment is time-varying as the other agents also  adjust their policies. Policy based methods aim to achieve the  optimal performance when other agents play arbitrarily by minimizing the 
   \textit{(external) regret}, that is, find a sequence of actions that perform nearly as well as  the  optimal fixed policy   in hindsight.
 An algorithm with negligible average overall    regret is  called \textit{no-regret} or \textit{Hannan consistent}  \cite{hannan1957approximation}.
 Any Hannan consistent algorithm is  known to have the following two desired properties in repeated normal-form games. First, when other agents adopt stationary policies, the  time-average policy constructed by the  algorithm converges to the best response policy (against the ones used by the other agents). Second, more interestingly, in two-player zero-sum games, when both players adopt Hannan consistent algorithms and both their  average overall regrets are no more than $\epsilon$,    their time-average policies constitute a $2\epsilon$-approximate Nash equilibrium \cite{blum2007learning}. 
  Thus, 
  any Hannan consistent single-agent reinforcement learning algorithm can be applied to find the Nash equilibria of  
   two-player zero-sum games via \emph{self-play}. Most of    these methods belong to one of  the following two families: 
  fictitious play \cite{brown1951iterative, robinson1951iterative}, and counterfactual regret minimization \cite{zinkevich2008regret}, which will be summarized below.

  Fictitious play  is a classic algorithm studied in game 
theory, where  the players  play the game repeatedly and each player adopts a policy that  best responds to the average policy of the other agents. 
This method was originally proposed for solving 
 \textit{normal-form games}, which are a simplification of the Markov  games defined in Definition \ref{def:Markov_Game} with $\cS  $  being a singleton and $\gamma = 0$.  
 In particular, for  any joint policy $\pi \in \Delta(\cA)$    of the  $N$ agents, we let $\pi^{-i}$ be the marginal policy of all players except player $i$. For any $t\geq 1$, suppose the agents have played $\{ a_\tau \colon.1\leq  \tau   \leq t \} $ in the first $t$ stages. We define $x_{t }$ as the empirical distribution of  $\{ a_\tau \colon.1\leq  \tau   \leq t \} $, i.e., $x_t(a) = t^{-1} \sum_{\tau =1}^t\mathbbm{1} \{ a_t = a \} $ for any $a \in \cA$.  Then, in the $t$-th stage, each agent $i$ takes action $a_t^i \in \cA^i $ according to the best response policy against $x_t^{-i} $.  In other words, each agent plays the best counter policy against  the policy of the other agents inferred from  history data. 
 Here, for any $\epsilon >0$ and any $\pi \in \Delta(\cA)$, we denote by $\texttt{Br}_{\epsilon} (\pi^{-i})$ the $\epsilon$-best response policy of player $i$, which satisfies 
 \#\label{eq:br_policy}
 R^i \bigl  ( \texttt{Br}_{\epsilon} (\pi^{-i}), \pi^{-i} \bigr ) \geq \sup_{\mu\in \Delta(\cA^i)} R^i(\mu, \pi^{-i})  -\epsilon.
 \#
 Moreover, we define $\texttt{Br}_{\epsilon} (\pi)$ as the joint policy $( \texttt{Br}_{\epsilon} (\pi^{-1}) , \ldots,  \texttt{Br}_{\epsilon} (\pi^{-N}) ) \in \Delta (\cA) $ and suppress the subscript $\epsilon$ in $\texttt{Br}_{\epsilon}$ if $\epsilon = 0$.
By this notation, regarding each $a \in \cA$ as a vertex of $\Delta( \cA)$, we can equivalently write the fictitious process as 
 \#\label{eq:DFP}
 x_t - x_{t-1} = (1/ t)  \cdot (a_t - x_{t-1} ), \qquad \text{where}\qquad a_t \sim \texttt{Br}( x_{t-1}  )  . 
 \#
 As $t\to\infty$, the updates  in  \eqref{eq:DFP}  can be approximately characterized by  a differential inclusion \cite{benaim2005stochastic} 
 \#\label{eq:CFP}
 \frac{\ud x(t)  }{\ud t} \in  \texttt{Br}( x(t) ) - x(t),
 \#
 which is known as the continuous-time fictitious play. Although it is well known that  the discrete-time fictitious play  in \eqref{eq:DFP} is not Hannan consistent \cite{hart2001general, young1993evolution}, it is shown in 
 \cite{monderer1997belief, viossat2013no} that the continuous-time fictitious play  in  \eqref{eq:CFP}   is Hannan consistent. 
 Moreover, using tools from stochastic approximation \cite{kushner2003stochastic, hart2001general},  
 various modifications of  discrete-time  fictitious play based on techniques such as  smoothing or stochastic perturbations have been  shown to converge to  the  continuous-time fictitious play   and are thus  Hannan consistent
 \cite{fudenberg1995consistency, hofbauer2002global,  leslie2006generalised, benaim2013consistency, li2018sampled}. As a result,  applying these methods with self-play provably finds a Nash equilibrium of a two-player zero-sum normal form game.

 Furthermore, fictitious play methods have also been extended to RL settings without the model knowledge.  Specifically,  using  sequence-form representation,
 \cite{heinrich2015fictitious}  has proposed the first fictitious play algorithm  for extensive-form games which is realization-equivalent to the \textit{generalized weakened fictitious play} \cite{leslie2006generalised} for normal-form games. 
 The pivotal insight is that a convex combination of normal-form policies can be written as a weighted convex combination of behavioral policies using realization  probabilities. 
 Specifically,  recall that  the set of information states of agent $i$ was denoted by $\cS^i$.  
 When the game has perfect-recall, each $s^i \in \cS^i$ uniquely defines a sequence $\sigma_{s^i}$ of actions played by agent $i$ for reaching state $s^i$.  Then any behavioral policy $\pi^i$ of agent $i$ induces a \textit{realization probability} $\texttt{Rp}(\pi^i;  \cdot  )$  for each  sequence $\sigma$ of agent $i$, which is defined by $\texttt{Rp}(\pi^i;  \sigma ) = \prod_{(\sigma_{s'},  a) \sqsubseteq \sigma } \pi^i( a \given s') $, where the product is taken over all $s' \in \cS^i$ and $a\in \cA^i$ such that   $(\sigma_{s'} , a) $ is a subsequence of $\sigma$. 
 Using the notation of realization probability, for any two behavioral policies  $\pi$ and $\tilde \pi$ of agent $i$,  the sum 
 \#\label{eq:mixture}
   \frac{\lambda \cdot \texttt{Rp} (\pi, \sigma_{s^i})  \cdot \pi( \cdot \given s^i) }  { \lambda \cdot \texttt{Rp} (\pi, \sigma_{s^i})  + (1- \lambda) \cdot \texttt{Rp} ( \tilde \pi  , \sigma_{s^i})}+  \frac{(1- \lambda) \cdot \texttt{Rp} ( \tilde \pi , \sigma_{s^i}) \cdot  \tilde \pi (\cdot \given s^i )}  { \lambda \cdot \texttt{Rp} (\pi, \sigma_{s^i})  + (1- \lambda) \cdot \texttt{Rp} ( \tilde \pi , \sigma_{s^i})}  , \qquad \forall s^i \in \cS^i, 
 \#
is the mixture policy of  $\pi$ and $\tilde \pi$ with weights $\lambda \in (0,1)$ and $1-\lambda$, respectively. 
Then, combining \eqref{eq:br_policy} and  \eqref{eq:mixture},  the  fictitious play algorithm in \cite{heinrich2015fictitious}   computes a sequence of policies $\{ \pi_t \}_{t\geq 1}$.  
In particular, 
in the $t$-th iteration,  any agent $i$ first computes the $\epsilon_{t+1}$-best response policy $\tilde \pi_{t+1} ^i \in  \tt{Br}_{\epsilon_{t+1}}(\pi^{-i}_t)$ and then  constructs $\pi^i_{t+1}$ as  the mixture policy of $\pi^i_{t}$ and $\tilde \pi_{t+1}$ with weights $1- \alpha_{t+1}$ and $\alpha_{t+1}$, respectively.  Here,  both $\epsilon_t$  and $\alpha_t $ are taken to converge to zero as $t$ goes to infinity, and we further have $\sum_{t\geq 1} \alpha_t = \infty$. 
 We note, however, that although such a method provably converges to a Nash equilibrium of a zero-sum game via self-play, it suffers from the curse of dimensionality due to the need to iterate all states of the game in each iteration. For computational efficiency, \cite{heinrich2015fictitious} has also proposed a data-drive fictitious self-play framework where the best-response is computed via  fitted Q-iteration \cite{ernst2005tree, munos2007performance}  for the single-agent RL problem,  with the policy mixture being  learned through   supervised learning. This framework was later  adopted by  \cite{heinrich2014self, heinrich2016deep, kawamura2017neural,  zhang2019monte} to incorporate other single RL methods such as deep Q-network \cite{mnih2015human} and Monte-Carlo tree search  \cite{coulom2006efficient, kocsis2006bandit, browne2012survey}. Moreover, in a more recent work, \cite{perolat2018actor} has proposed a smooth fictitious  play algorithm \cite{fudenberg1995consistency} for zero-sum multi-stage games with simultaneous moves (a special case of zero-sum stochastic games). 
 Their algorithm combines the 
 actor-critic  framework \cite{konda2000actor}  with fictitious self-play, and infers the  opponent's policy implicitly via policy evaluation. Specifically, when the two players 
 adopt a joint policy $\pi = (\pi^1, \pi^2)$,  
 from the perspective of player $1$,  it infers $\pi^2$ implicitly by estimating $  \overbar Q _{\pi^1, \pi^2} $ via temporal-difference learning  \cite{sutton1987temporal}, where $ \overbar Q_{\pi^1, \pi^2} \colon \cS \times \cA^1 \rightarrow \RR$ is defined as 
 \$
 \overbar Q_{\pi^1, \pi^2} (s, a^1 ) = \EE\bigg[\sum_{t\geq 0}\gamma^t R^1(s_t,a_t,s_{t+1})\bigggiven s_0 = s, a_0^1 = a^1, a_0^2 \sim \pi^2 (\cdot \given  s) , a_t\sim \pi(\cdot\given s_t), \forall t\geq 1  \bigg] ,
 \$
 which is 
  the action-value function of player $1$ marginalized by $\pi^2$. Besides, the best response policy is obtained by taking the soft-greedy policy with respect to $\overbar Q_{\pi^1, \pi^2} $, i.e.,
  \#\label{eq:soft_greedy}
  \pi^1 (a^1  \given s)  \leftarrow \frac{\exp[  \eta ^{-1} \cdot \overbar Q_{\pi^1, \pi^2} (s,a^1 ) ] }{\sum_{a^1 \in \cA^1} \exp[  \eta^{-1} \cdot \overbar Q_{\pi^1, \pi^2} (s,a^1 ) ]  } ,
  \#
  where $\eta>0$ is the  smoothing parameter.  Finally, the algorithm is obtained by performing both policy evaluation and policy update in  \eqref{eq:soft_greedy} simultaneously using two-timescale updates \cite{borkar2008stochastic,kushner2003stochastic}, which ensure that the   policy updates, when using self-play, can be characterized by an ordinary differential equation whose asymptotically stable solution is 
  a smooth Nash equilibrium of the game.

 Another  family of popular  policy-based methods is based on the idea of \textit{counterfactural regret minimization} (CFR),   first proposed in \cite{zinkevich2008regret}, which has been a breakthrough in the effort to solve large-scale extensive-form games. 
Moreover, from a theoretical perspective, compared with fictitious play algorithms whose convergence is analyzed asymptotically via stochastic approximation, explicit regret bounds can be established using tools from online learning \cite{cesa2006prediction}, which yield rates of convergence to the  Nash equilibrium. 
  Specifically, 
  when $N$ agents play the extensive-form game for  $T$ rounds with  $\{ \pi_t \colon 1\leq t \leq T\}$, the \textit{regret} of player $i$   is defined~as 
  \#\label{eq:regret}
  \texttt{Reg}_T^i = \max_{\pi^i } \sum_{t = 1}^T \bigl [ R^i (\pi^i, \pi^{-i} _t  ) - R^i(\pi_t^i, \pi_t^{-i} ) \bigr ] ,
  \#
  where the maximum is taken over all possible policies of player $i$.
  In the following, before we define the notion of counterfactual regret, we first introduce a few notations. 
 Recall that we had defined the reach probability $\eta_{\pi} (h)$ in \eqref{eq:reach_prod}, which can be factorized into the product of each agent's contribution. That is, for each $i \in \cU \cup \{c\}$, we can group the probability   terms involving $\pi^{i}$ into $ \eta_{\pi}^i (h) $ and write 
 $
\eta_{\pi}(h) = \prod_{i \in \cN \cup \{c\} }  \eta_{\pi}^i (h) =  \eta_{\pi}^i (h) \cdot  \eta_{\pi}^{-i} (h).
$
Moreover, for any two histories $h , h' \in \cH$ satisfying $ h\sqsubseteq h'$, 
we define the \textit{conditional reach probability} 
$\eta_{\pi} ( h' \given h) $ as $ \eta _{\pi}(h') /  \eta_{\pi}(h)$ and define $\eta_{\pi}^i( h' \given h) $ similarly.
For any $s \in \cS^i $ and any $a \in \cA(s)$, we define $\cZ (s,a) = \{(h,z) \in \cH \times \cZ\given h \in s , ha \sqsubseteq z \}$, which contains all possible pairs of  history in   information state $s$ and    terminal history  after taking action $a$ at $s$.
Then, the \textit{counterfactual value function} is defined as 
\#
Q^i_{\texttt{CF}} (\pi, s, a)  & = \sum_{(h, z) \in \cZ (s,a) } \eta_{\pi}^{-i} (h) \cdot \eta_{\pi} ( z \given ha)  \cdot R ^{i} (z), \label{eq:cf_value_fun1}
 \#
which is the expected utility obtained by agent $i$ given that it has played to reached state~$s$.  
We also define  
$
 V^i_{\texttt{CF}} (\pi, s )  =  \sum_{a \in \cA(s) } Q^i_{\texttt{CF}} (\pi, s, a) \cdot \pi^i (a \given s)$.
  Then the difference between $Q^i_{\texttt{CF}} (\pi, s, a)  $ and $V^i_{\texttt{CF}} (\pi, s ) $ can be viewed as the value of action $a $ at information state $s \in \cS^i$, and  \textit{counterfactual regret} of agent $i$ at state $s$   is defined as 
\#\label{eq:immidiate_regret}
\texttt{Reg} _{T}^i (s ) = \max_{a \in \cA(s) } \sum_{i =1}^T \bigl [Q^i_{\texttt{CF}} (\pi_t, s, a)  - V^i_{\texttt{CF}} (\pi_t, s ) \bigr ], \qquad \forall s \in \cS^i. 
\#
Moreover, as shown in Theorem 3 of  \cite{zinkevich2008regret}, counterfactual regrets defined in \eqref{eq:immidiate_regret} provide an upper bound for the total regret in \eqref{eq:regret}: 
\#\label{eq:regret_upper}
\texttt{Reg}_T^i  \leq \sum_{s \in \cS^i } \texttt{Reg} _{T} ^{i , + }(s ) ,
\# 
where we let $x ^{+ }$ denote $\max \{ x, 0 \}$ for any $x \in \RR$. 
This bound lays the foundation of  \textit{counterfactual regret minimization} algorithms. Specifically, to minimize the total regret in \eqref{eq:regret}, it suffices to minimize the counterfactual regret for each information state, which can be obtained by any online learning algorithm, such as  EXP3   \cite{auer2002nonstochastic}, Hedge \cite{vovk1990aggregating, littlestone1994weighted,
 freund1999adaptive}, and regret matching \cite{hart2000simple}. 
All these methods ensure that the counterfactual regret is $   \cO(\sqrt{T})$ for all $s \in \cS^i$, which leads to an  $\cO( \sqrt{T})$ upper bound of the total regret. Thus,  applying CFR-type methods with self-play to a zero-sum two-play extensive-form game, the average policy is   an $\cO(\sqrt{1/T})$-approximate Nash equilibrium after $T$ steps.
In particular, 
 the vanilla  CFR algorithm updates the policies via regret matching, which yields that   
$  
 \texttt{Reg} _{T}^i (s )  \leq R_{\max}^i \cdot \sqrt{ A^i   \cdot T} 
$
for all $s\in \cS^i$, where we have introduced   
$$R_{\max}^i = \max_{z \in \cZ} R^i(z) - \min_{z \in \cZ} R^i(z), \qquad A_i = \max_{h \colon \tau(h) = i} | \cA(h ) |. $$
Thus, by  \eqref{eq:regret_upper},  the total regret of agent $i$ is bounded by $R_{\max}^i \cdot | \cS^i | \cdot  \sqrt{ A^i   \cdot T}$.  

One drawback of vanilla CFR is that the entire game tree needs to be traversed in each iteration, which can be computationally prohibitive. A number of  CFR variants have been proposed since   
the  pioneering work \cite{zinkevich2008regret} for  improving   computational efficiency. For example,  \cite{lanctot2009monte, burch2012efficient,gibson2012generalized, johanson2012efficient, lisy2015online,schmid2019variance} combine CFR with Monte-Carlo sampling;   \cite{waugh2015solving,morrill2016using, brown2019deep} propose to   estimate the counterfactual value functions via regression;  \cite{brown2015regret,brown2017dynamic, brown2017reduced}   improve the computational efficiency by  pruning  suboptimal paths in the game tree;    \cite{tammelin2014solving,tammelin2015solving, burch2019revisiting}  analyze the performance of a modification named  $\text{CFR}^{+}$, and \cite{zhou2018lazy} proposes lazy updates  with a near-optimal regret upper bound.   
 
 Furthermore, it has been  shown recently in \cite{srinivasan2018actor} that CFR is closely related to policy gradient methods. To see this, for any joint policy $\pi$ and any $i \in \cN$,
 we define the action-value function of agent $i$, denoted by $ Q_{\pi}^i $,  as 
 \#\label{eq:Q_efg}
 Q_{\pi}^i (s, a) =  \frac{1} {\eta_{\pi} (s) }  \cdot  \sum_{(h, z) \in \cZ (s,a) } \eta_{\pi}  (h)   \cdot \eta_{\pi} ( z \given ha)  \cdot R ^{i} (z),  \qquad \forall s \in \cS^i, \forall a \in \cA(s).
 \#
 That is,  $Q_{\pi}^i (s, a)$ 
 is the expected utility of agent~$i$ when the agents follow policy $\pi$ and agent $i$ takes action $a$ at information state $s \in \cS^i$, conditioning on $s $ being  reached. It has been  shown   in \cite{srinivasan2018actor} that the $Q^i_{\texttt{CF}}$  in 
 \eqref{eq:cf_value_fun1} is connected with $Q_{\pi}^i $ in \eqref{eq:Q_efg} via 
  $Q^i_{\texttt{CF}} (\pi, s, a)  =Q_{\pi}^i (s,a) \cdot  [ \sum_{h \in s } \eta^{-i} _{\pi} (h) ]   $. 
Moreover, in the tabular setting where we regard the joint policy $\pi$ as a table $\{ \pi^i (a\given s) \colon s \in \cS^i, a \in \cA(s) \}$, for any $s \in \cS^i$ and any $a \in \cA(s)$,  the policy gradient of $R^i (\pi) $ can be written~as 
  \$
\frac{\partial R^i (\pi) } { \partial _{\pi^i(a\given s) } }  =  \eta_{\pi}(s) \cdot  Q_{\pi}^i (s, a) = \eta_{\pi}^i (s) \cdot Q^i_{\texttt{CF}} (\pi, s, a) , \qquad \forall s \in \cS^i, \forall a \in \cA(s). 
  \$
  As a result, the advantage actor-critic  (A2C) algorithm \cite{konda2000actor}   is equivalent to a particular  CFR  algorithm, where the policy update rule is specified by the \textit{generalized infinitesimal gradient ascent} algorithm \cite{zinkevich2003online}.
 Thus,  \cite{srinivasan2018actor} proves that the regret of the  tabular A2C algorithm    is bounded by 
$
  | \cS^i | \cdot [  1 + A^i \cdot (R^i_{\max} )^2 ] \cdot  \sqrt{T}.     $    
  Following this work,  \cite{omidshafiei2019neural} shows that A2C where the policy is tabular and  is parametrized by a softmax function  is equivalent to CFR that uses  Hedge to update the policy. 
  Moreover,  \cite{lockhart2019computing} proposes a policy optimization method known as \textit{exploitability descent}, where the policy is updated using actor-critic, assuming the opponent plays the best counter-policy. This method is equivalent to the CFR-BR algorithm \cite{johanson2012finding} with Hedge.  
Thus,   \cite{srinivasan2018actor,omidshafiei2019neural, lockhart2019computing} show that actor-critic and policy gradient  methods for MARL can be formulated as  CFR methods and thus convergence to a Nash equilibrium of a zero-sum extensive-form game is guaranteed.
         

  In addition, besides   fictitious play and CFR methods introduced above,  multiple policy optimization methods have been  proposed for special classes of two-player zero-sum stochastic games  or extensive form games. For example, Monte-Carlo tree search methods
   have been  proposed  for perfect-information extensive games with simultaneous moves. 
It has been shown in \cite{schaeffer2009comparing} that  the MCTS  methods with UCB-type action selection rules, introduced in \S\ref{sec:value_comp}, fail to converge to a Nash equilibrium in simultaneous-move games, as UCB does not take into consideration the possibly adversarial moves of the opponent. 
To remedy this issue, 
   \cite{lanctot2013monte,lisy2013convergence, tak2014monte, kovavrik2018analysis} 
 have proposed to adopt stochastic policies and using  Hannan consistent methods such as EXP3   \cite{auer2002nonstochastic} and regret matching \cite{hart2000simple} to update the policies. With self-play, \cite{lisy2013convergence} shows that the average policy obtained by MCTS with any $\epsilon$-Hannan consistent policy update method  converges to an $\cO(D^2 \cdot \epsilon)$-Nash equilibrium, where $D$ is the maximal depth.

 Finally, there are surging interests  in investigating   
 policy gradient-based methods in \emph{continuous} games, i.e., the games with continuous state-action spaces. With policy parameterization, finding the NE of  zero-sum Markov games becomes   a nonconvex-nonconcave saddle-point  problem in general \cite{mazumdar2018convergence,mazumdar2019finding,zhang2019policyb,bu2019global}. This hardness is inherent, even in the simplest linear quadratic setting with linear function approximation \cite{zhang2019policyb,bu2019global}. As a consequence,     most of the convergence results are \emph{local}  \cite{mescheder2017numerics,mazumdar2018convergence,adolphs2018local,daskalakis2018limit,mertikopoulos2019optimistic,fiez2019convergence,mazumdar2019finding,jin2019minmax}, in the sense that they address  the convergence behavior around  local NE points. Still, it has been shown that the vanilla gradient-descent-ascent (GDA) update, which is equivalent to the policy gradient update in MARL, fails to converge to local NEs, for either the non-convergent behaviors such as limit cycling  \cite{mescheder2017numerics,daskalakis2018limit,balduzzi2018mechanics,mertikopoulos2019optimistic}, or the existence of non-Nash stable limit points for the GDA dynamics \cite{adolphs2018local,mazumdar2019finding}. Consensus optimization \cite{mescheder2017numerics},   symplectic gradient adjustment \cite{balduzzi2018mechanics}, and extragradient method   \cite{mertikopoulos2019optimistic} have been advocated to mitigate the oscillatory behaviors around the equilibria; while \cite{adolphs2018local,mazumdar2019finding} exploit the curvature information so that all the stable limit points of the proposed updates are local NEs. 
Going beyond Nash equilibria, \cite{jin2019minmax,fiez2019convergence} consider gradient-based learning for \emph{Stackelberg equilibria}, which correspond to only the one-sided equilibrium solution in zero-sum games, i.e., either minimax or maximin, as the order of which player acts first is vital in nonconvex-nonconcave problems.  
\cite{jin2019minmax} introduces the  concept of \emph{local minimax} point as the solution, and shows that GDA converges to  local minimax points under mild conditions.  
\cite{fiez2019convergence}  proposes a two-timescale algorithm where the follower uses a gradient-play update rule, instead of an exact best response strategy, which has been shown to converge to the Stackelberg equilibria. 
 Under a stronger assumption of \emph{gradient dominance}, \cite{sanjabi2018solving,nouiehed2019solving} have shown that nested gradient descent methods converge to the stationary points of the outer-loop, i.e., minimax, problem at a sublinear rate. 
 





We note that these convergence results have been developed for \emph{general}  continuous  games with \emph{agnostic} cost/reward functions, meaning that the functions may have various forms, so long as they are \emph{differentiable}, sometimes even \emph{(Lipschitz) smooth}, w.r.t.  each agent's policy parameter. For MARL, this is equivalent to requiring   differentiability/smoothness of the long-term \emph{return}, which relies on the properties of the  game, as well as of  the policy parameterization. Such an assumption is generally very restrictive. For example,   the Lipschitz smoothness assumption fails to hold globally for LQ games \cite{zhang2019policyb,mazumdar2019policy,bu2019global}, a special type of MGs.  Fortunately, thanks to the special structure of the LQ setting, \cite{zhang2019policyb} has proposed several   projected nested policy gradient methods that are guaranteed to have \emph{global} convergence to the NE, with convergence rates established.  This appears to be the first-of-its-kind result in MARL.  The results have then been improved by the techniques in a subsequent work of the authors  \cite{zhang2019policy}, which can  remove the projection step in the updates, for a more general class of such games. 
Very recently, \cite{bu2019global} also improves the results in  \cite{zhang2019policyb} independently, with different techniques.  


\subsection{Mixed Setting}\label{subsec:mixed_setting}

In stark contrast with the fully  collaborative  and fully competitive settings, the mixed setting is notoriously challenging and thus  rather less well understood. 
Even in the simplest case of a  two-player general sum normal-form game, finding a  Nash  equilibrium is PPAD-complete \cite{chen2009settling}.   
Moreover, \cite{zinkevich2006cyclic} has proved  that value-iteration methods fail to find stationary Nash or correlated equilibria for general-sum Markov games.  Recently,   it is shown that vanilla  policy-gradient   methods avoid a non-negligible subset of   Nash equilibria in 
  general-sum continuous games \cite{mazumdar2018convergence},  including the  LQ general-sum games \cite{mazumdar2019policy}.   
Thus,  additional structures on either the games or the algorithms   
need to be exploited, to ascertain provably convergent MARL in the mixed setting.

\vspace{6pt}
\noindent{\bf Value-Based Methods}
\vspace{3pt} 
 
Under relatively stringent assumptions, 
several value-based methods that extend 
Q-learning \cite{watkins1992q} 
 to the mixed setting are guaranteed to find an equilibrium.  
In particular, \cite{hu2003nash}  has proposed the Nash-Q learning  algorithm for general-sum Markov games, where  one maintains $N$  action-value functions $ Q_{\cN} = (Q^1, \ldots, Q^N) \colon    \cS \times \cA \rightarrow \RR^N  $  for all $N$ agents,  which are  updated 
 using sample-based estimator of a Bellman operator. Specifically,  letting $ R_{\cN} = ( R^1, \ldots, R^N)$ denote the reward functions of the agents,  Nash-Q uses the following Bellman operator: 
 \# \label{eq:nash_bellman}
   ( \cT^* Q_{\cN} )(s,a )  = \EE _{  s' \sim \cP(\cdot \given s, a )} \big  \{ R_{\cN}  (s, a , s' ) + \gamma \cdot \texttt{Nash}\bigl[  Q_{\cN} (s', \cdot ) \bigr ] \bigr \} , \quad \forall (s , a) \in \cS\times \cA,
   \#
where   $\texttt{Nash}[  Q_{\cN} (s', \cdot )   ]$  is the objective value of the Nash equilibrium of the stage game with rewards $\{ Q_{\cN} (s', a ) \}_{a \in \cA}   $. For zero-sum games, we have $Q^1 = -Q^2$ and thus the Bellman operator defined in \eqref{eq:nash_bellman} is equivalent to the one in  \eqref{eq:minimax_Q} used by minimax-Q learning \cite{littman1994markov}.  Moreover, \cite{hu2003nash} establishes  convergence to Nash equilibrium under the    restrictive assumption that  $\texttt{Nash}[  Q_{\cN} (s', \cdot )   ]$ in each iteration of the algorithm has unique Nash equilibrium.  
In addition, \cite{littman2001friend} has proposed the 
   Friend-or-Foe Q-learning   algorithm where each agent views the  other agent as either a ``friend'' or a ``foe''. In this case, $\texttt{Nash}[  Q_{\cN} (s', \cdot )   ]$ can be efficiently computed via linear programming.  This algorithm can be viewed as a  generalization of minimax-Q learning, and Nash equilibrium convergence is guaranteed for two-player zero-sum games and coordination games with a unique equilibrium. Furthermore,  
 \cite{greenwald2003correlated} has proposed  correlated Q-learning, which replaces $\texttt{Nash}[  Q_{\cN} (s', \cdot )   ]$ in \eqref{eq:nash_bellman} by computing a correlated equilibrium \cite{aumann1974subjectivity}, a more general equilibrium concept than Nash equilibrium. In a recent work, \cite{pmlr-v54-perolat17a}  has proposed a batch RL method to find an approximate  Nash equilibrium  via    Bellman residue minimization \cite{maillard2010finite}. They have proved that the global minimizer of the empirical Bellman residue is an approximate  Nash equilibrium, followed by the error propagation analysis for the algorithm.  
Also in the batch RL regime, \cite{zhang2018finite} has considered a simplified mixed setting for decentralized MARL: two teams of cooperative networked agents compete in a zero-sum Markov game. 
A decentralized  variant of FQI, where the agents within one team cooperate to solve \eqref{equ:FQI_coop} while the two teams essentially solve \eqref{eq:FQI}, is proposed. 
Finite-sample error bounds have  then been established for the proposed algorithm.

To address  the scalability issue, independent learning is preferred, which, however, fails to converge in general  \cite{tan1993multi}.      \cite{arslan2017decentralized} has proposed  \emph{decentralized Q-learning}, a two
timescale modification of Q-learning, that is guaranteed to converge to the equilibrium  for \emph{weakly acyclic Markov games} almost surely. Each agent therein only  observes local action and reward, and neither observes nor  keeps track of others' actions.  
All agents are instructed to use the same  stationary \emph{baseline policy} for many consecutive stages, named \emph{exploration phase}. At the end of the \emph{exploration phase}, all agents are \emph{synchronized} to update their baseline policies, which makes the environment stationary for long enough, and enables the convergence of Q-learning based methods.  
Note that these algorithms can also be applied  to the cooperative setting, as these games include Markov teams as a special case.

 \vspace{6pt}
\noindent{\bf Policy-Based Methods}
\vspace{3pt} 
  
For continuous games, 
due to the general negative  results  therein,  \cite{mazumdar2018convergence} introduces  a new class of games, \emph{Morse-Smale games}, for which the gradient dynamics correspond to gradient-like flows. Then, definitive statements on  almost sure convergence of PG methods  to either limit cycles, Nash equilibria, or non-Nash fixed points can be made, using tools from  dynamical systems theory.  
  Moreover,  \cite{balduzzi2018mechanics,letcher2019differentiable} have studied   the second-order structure of game dynamics, by decomposing  it into two components.The first one, named symmetric component,  relates to potential games, which yields gradient descent on some implicit function; the second one, named antisymmetric component, relates  to  \emph{Hamiltonian games} that follows some conservation law,  motivated by classical mechanical systems analysis. The fact that gradient descent converges to the Nash equilibrium of both types of games motivates the development of  the Symplectic Gradient Adjustment  algorithm that finds  \emph{stable fixed points} of the game, which constitute all local Nash equilibria for zero-sum games, and only 
  a subset of local NE for general-sum games. 
\cite{chasnov2019convergence} provides finite-time local convergence guarantees to  a neighborhood of a \emph{stable} local Nash equilibrium of continuous games, in both deterministic setting, with exact PG, and stochastic setting, with unbiased PG estimates. Additionally, \cite{chasnov2019convergence} has also explored the effects of \emph{non-uniform} learning rates on the learning dynamics and convergence rates. 
\cite{fiez2019convergence} has also considered \emph{general-sum} Stackelberg games, and shown that the same  two-timescale algorithm update as in the zero-sum case now converges almost surely to  the stable attractors only. It has also established  finite-time performance  for local convergence to a neighborhood of a stable Stackelberg equilibrium.  
In complete analogy to the zero-sum class, these convergence results for continuous games do not apply to MARL in Markov games directly, as they are  built upon the differentiability/smoothness of the long-term return, which may not hold for  general MGs, for example, LQ games \cite{mazumdar2019policy}.  Moreover, most of these convergence results are local instead of global.

Other than continuous games,  
the policy-based methods summarized  in \S\ref{sec:policy_comp} can also be applied to the mixed setting via self-play. 
The validity of such an approach is based a fundamental connection between game theory and online learning -- If the external regret of each agent is no more than $\epsilon$, then their  average policies constitute an $\epsilon$-approximate
 \textit{coarse correlated equilibrium}  \cite{hart2000simple,hart2001general, hart2003uncoupled} of the general-sum normal-form games. Thus, although in general we are unable to find a Nash equilibrium, policy optimization with self-play   guarantees to  find a coarse correlated equilibrium in these normal-form games.

\vspace{6pt}
\noindent{\bf Mean-Field Regime}
\vspace{3pt}

The scalability issue in the non-cooperative setting can also be alleviated in the mean-field  regime, as the cooperative setting discussed in \S\ref{subsubsec:homo_agents}. 
For general-sum games, 
\cite{pmlr-v80-yang18d} has proposed a modification of the Nash-Q learning algorithm where the actions of other agents are approximated by their empirical average. 
That is, the action value function of each agent $i$ is parametrized by $Q^i (s, a^i, \mu_{a^{-i}} )$, where $\mu_{a^{-i}}$ is the empirical distribution of  $\{a_j \colon j \neq i \}$.
 Asymptotic convergence of this mean-field  Nash-Q learning algorithm has also been established.

Besides, most mean-field RL algorithms are focused on addressing the   \textit{mean-field game} model.  
In mean-field games, each agent $i$ has a local state $s^i \in \cS $ and a local action $a^i \in \cA$, 
and  the interaction among other agents is  captured  by an aggregated effect $\mu$, also known as the \emph{mean-field term}, which is a functional of the empirical distribution of the local states and actions of the agents.  
Specifically, at the $t$-th time step, when  agent $i$ takes action $a_t^i$ at state $s_t^i$ and the mean-field term is $\mu_t$, it receives an immediate reward $R(s_t^i, a_t^i, \mu_t)$ and its local state evolves into $  s_{t+1}^i \sim \cP(\cdot \given s_t^i, a_t^i, \mu_t) \in \Delta(\cS)$. Thus, from the perspective of agent $i$,  instead of participating in a multi-agent game, it is faced with a time-varying MDP parameterized by the sequence of mean-field terms $\{ \mu_t \}_{t\geq 0}$, which in turn is determined by the states and actions of all agents. 
The solution concept in MFGs is the   \emph{mean-field equilibrium}, which is a sequence of \emph{pairs} of policy and mean-field terms
$\{\pi^*_t, \mu^*_t \}_{t\geq 0}$ that satisfy the following two conditions: (1) $\pi^* = \{\pi^*_{t} \}_{t\geq 0}$ is the optimal policy for the time-varying MDP specified by $\mu^* = \{\mu^*_t \}_{t\geq 0}$, and (2)  $\mu^*$ is generated when each agent follows policy $\pi^*$.  The existence of the mean-field equilibrium for discrete-time MFGs has been  studied in  
\cite{saldi2018markov, saldi2019approximate,saldi2019discrete,saldi2018discrete}
and their constructive  proofs exhibit that the mean-field equilibrium can be obtained via a fixed-point iteration. Specifically, one can construct a sequence of policies and mean-field terms $\{\pi^{(i)} \}_{i\geq 0}$ and $\{ \mu^{(i)}\} _{i\geq 0}$ such that $\{ \pi^{(i)}\}_{t\geq 0}$ solves the time-varying MDP specified by $\mu^{(i)}
$, and $\mu^{(i+1)}$ is generated when all players adopt policy $\pi^{(i)}$. 
Following this agenda, various model-free RL methods are proposed for solving MFGs where $\{\pi^{(i)} \}_{i\geq 0}$ is approximately solved via single-agent RL such as Q-learning \cite{guo2019learning} and policy-based methods 
\cite{subramanian2019reinforcement, fu2019actor}, with  $\{ \mu^{(i)}\} _{i\geq 0}$ being  estimated via sampling.  
In addition,   \cite{hadikhanloo2019finite,elie2019approximate} recently propose   fictitious play updates for the mean-field state where we have $\mu^{(i+1)} = (1-\alpha^{(i)}) \cdot \mu^{(i)} + \alpha^{(i)}  \cdot \hat  \mu^{(i+1)} $, with $ \alpha^{(i)}$
 being the learning rate and $\hat  \mu^{(i+1)}$ being the mean-field term generated by policy $\pi^{(i)}$. Note that the aforementioned works focus on  the settings with either \emph{finite} horizon \cite{hadikhanloo2019finite,elie2019approximate} or \emph{stationary} mean-field equilibria  \cite{guo2019learning,subramanian2019reinforcement,fu2019actor}  only. Instead, recent works 
 \cite{anahtarci2019value,zaman2019approx}  consider possibly non-stationary mean-field equilibrium in infinite-horizon settings, and develop equilibrium computation algorithms that have laid  foundations for model-free RL algorithms.

\section{Application Highlights}\label{sec:applications}

\begin{figure*}[!t] 
	\centering
	\begin{tabular}{cccc}
		\hskip-10pt\includegraphics[width=0.24\textwidth]{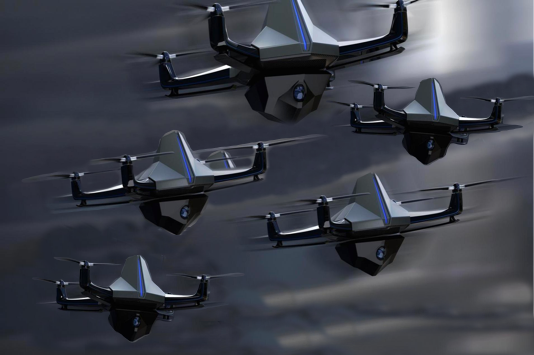}
		& 
		\hskip-5pt\includegraphics[width=0.24\textwidth]{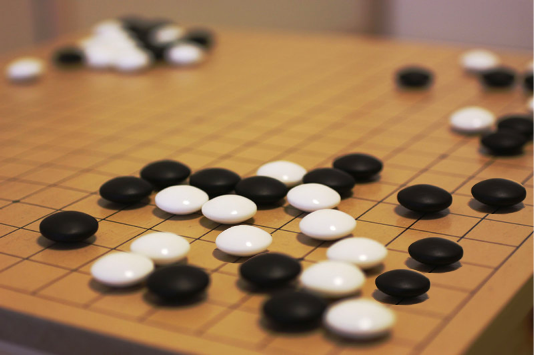}
		&   
		  \hskip-5pt\includegraphics[width=0.24\textwidth]{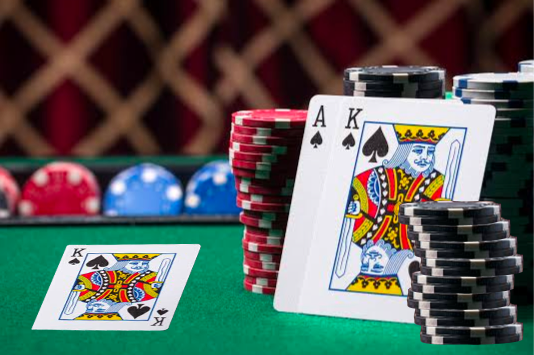}&   
		  \hskip-5pt\includegraphics[width=0.24\textwidth]{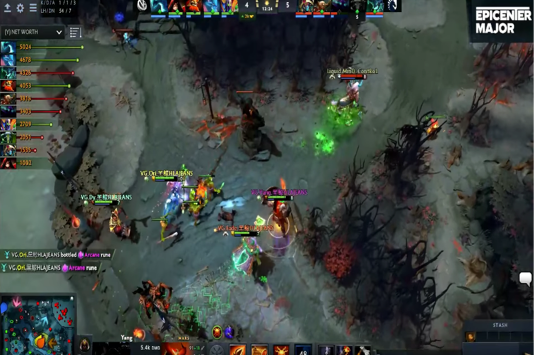}
\end{tabular}
	\caption{Four representative applications of recent successes of MARL: unmanned aerial vehicles, game of Go, Poker games, and team-battle video games. } 
	\label{fig:app} 
\end{figure*}

In this section, we briefly review the  recent empirical successes of 
MARL driven by the methods introduced in the previous section. 
In the following, we focus on the three MARL settings reviewed in  \S\ref{sec:algorithms} and highlight  four representative and practical applications in each setting, as illustrated in Figure \ref{fig:app}.

\subsection{Cooperative Setting} \label{sec:application_coop}

\noindent{\bf Unmanned Aerial Vehicles}
\vspace{3pt}  

One  prominent application of MARL is the control of practical multi-agent systems, most of which are cooperative and decentralized. 
Examples of the scenarios include  robot team navigation \cite{corke2005networked}, smart grid operation \cite{dall2013distributed}, and control of mobile sensor networks \cite{cortes2004coverage}. Here we choose  unmanned aerial vehicles (UAVs) \cite{yang2018keeping,pham2018cooperative,tovzivcka2018application,shamsoshoara2019distributed,cui2019application,qie2019joint}, a recently surging application scenario of multi-agent autonomous systems, as one representative example. 
 Specifically, a team of  UAVs are deployed to accomplish a cooperation task, usually without the coordination of any central controller, i.e., in a decentralized fashion. Each UAV is  normally equipped with communication devices, so that they can exchange information with some of  their  teammates, provided that they are  inside its sensing and coverage range. As a consequence, this application naturally fits in the decentralized paradigm with networked agents we advocated in \S\ref{subsubsec:networked_paradigm}, which is also illustrated in Figure  \ref{fig:info_struc} (b). Due to the high-mobility of UAVs, the communication links among agents are indeed \emph{time-varying} and fragile, making (online)  cooperation extremely challenging. 
Various challenges thus arise in the  context of cooperative UAVs, some of which have recently been  addressed by MARL.

In \cite{yang2018keeping}, the UAVs' optimal links discovery and selection problem is considered. Each UAV $u\in\cU$, where $\cU$ is the set of all UAVs, has the  capability to perceive the local available channels and  select a connected link over a common channel shared by another agent $v\in\cU$. Each UAV $u$ has its local set of channels $\cC_u$ with $\cC_u\bigcap \cC_v\neq \emptyset$ for any $u,v$, and a connected link between two adjacent UAVs is built if they announce their messages  on the same channel simultaneously.  Each UAV's local state is whether the previous  message has been successfully sent,  and its action is to choose a pair $(v,ch_u)$, with $v\in\cT_u$ and $ch_u\in \cC_u$, where $\cT_u$ is the set of teammates that agent $u$ can reach. The availability of local channels $ch_u\in \cC_u$ is modeled  as probabilistic, and the reward $\cR^u$ is calculated by the number of messages that are  successfully sent. Essentially, the algorithm in \cite{yang2018keeping} is based on independent Q-learning \cite{tan1993multi}, but with two heuristics to improve the tractability and convergence performance: by \emph{fractional slicing}, it  treats each dimension (fraction) of the action space independently, and  estimates the actual Q-value by the average of that for all fractions; by \emph{mutual sampling}, it shares both state-action pairs and a mutual Q-function parameter. 
  \cite{pham2018cooperative} addresses the problem of \emph{field coverage}, where the UAVs aim to provide a full coverage of an unknown field, while minimizing the overlapping sections among their field of views.  Modeled as a Markov team, the overall state $s$ is the concatenation of all local states $s_i$, which are defined as its $3$-D position coordinates  in the environment.  Each agent chooses to either head different directions, or go up and down, yielding $6$ possible actions.  A multi-agent Q-learning over the \emph{joint} action space is developed, with linear function approximation.  
In contrast, 
  \cite{shamsoshoara2019distributed} focuses  on spectrum sharing among a network of UAVs. Under a remote sensing task, the UAVs are categorized into two clusters: the relaying ones that provide relay services and  the other ones that gain  spectrum access for the remaining ones, which perform the sensing task. Such a problem can be modeled as a \emph{deterministic} MMDP, which can thus be solved by distributed Q-learning proposed in  \cite{lauer2000algorithm}, with optimality guaranteed.  Moreover, \cite{qie2019joint} considers the problem of \emph{simultaneous} target-assignment and path-planning for multiple UAVs. In particular, a team of UAVs $U_i\in\textbf{U}$, with each $U_i$'s position at time $t$ given by $(x_i^U(t),y_i^U(t))$, aim to cover all the targets $T_j\in\textbf{T}$ without collision with the threat  areas $D_i\in\textbf{D}$, as well as with other UAVs. For each $U_i$, a path  $P_i$ is planned  as $P_i=\{(x_i^U(0),y_i^U(0),\cdots,x_i^U(n),y_i^U(n))\}$, and the length of $P_i$ is denoted by $d_i$. Thus, the goal is to minimize $\sum_{i}d_i$ while the collision-free constraints are satisfied. 
  By penalizing the collision in the reward function, such a problem can be characterized as one with a mixed MARL setting that contains both cooperative and competitive agents. Hence, the MADDPG algorithm proposed in \cite{lowe2017multi} is adopted, with centralized-learning-decentralized-execution.  
  Two other tasks that can be tackled by MARL include resource allocation in UAV-enabled communication networks, using Q-learning based method \cite{cui2019application}, aerial surveillance and base defense in UAV fleet control, using policy optimization method in a purely centralized fashion \cite{tovzivcka2018application}.

\vspace{6pt}
\noindent{\bf Learning to Communicate}
\vspace{3pt} 

Another application of cooperative MARL aims to foster  communication and coordination among a team of agents without explicit human supervision. 
Such a type of problems is usually formulated as a Dec-POMDP involving $N$ agents, which is similar to the Markov game introduced in Definition \ref{def:Markov_Game} except that
each agent cannot observe the state $s\in \cS$ and that each agent has the same reward function $\cR$. More specifically, we assume that 
  each agent $i \in \cN$ receives  observations from set $\cY^i$ via  a noisy observation channel $\cO^i \colon \cS \rightarrow \cP( \cY_i)$ such that agent $i$ observes a random variable $y^i \sim \cO^i (\cdot \given s)$ when the   environment is at state $s$.  Note that this model can be viewed as a POMDP when there is a central planner that collects the observations of each agent and decides the actions for each agent.
  Due to the noisy observation channels, 
  in such a model the agents need to communicate with each other so as to  better infer the underlying state and make decisions that maximize the expected return shared by all  agents. Let $\cN_t^{i} \subseteq \cN$  be the neighbors of agent $i$ at the $t$-th time step, that is, agent $i$ is able to receive a message $m_t^{j\rightarrow i}$ from any agent $j \in \cN_t^i$ at time $t$. Let $I_t^i $ denote  the information agent $i$ collects up to time $t$, which is defined as 
  \#\label{eq:information}
 I_t^i = \Bigl \{ \Big ( o_\ell ^i, \{ a_{\ell}^j \}_{j\in \cN } , \{  m_{\ell }^{j\rightarrow i} \}_{ j \in \cN_\ell^i }  \Bigr )  \colon \ell \in \{ 0, \ldots, t-1\} \Bigr \}  \bigcup \big \{ o_{t}^i,   \bigr \} ,
   \# 
   which contains its history collected in  previous time steps and the observation   received at time $t$. With the information $I_t^i$, agent $i$ takes an action $a_t^i \in \cA^i$ and also 
   broadcasts messages $m_t^{i\rightarrow j}$ to all agents $j$ with $i \in \cN_t^j$. That is,  the policy $\pi_t^i$ of agent $i$ is a mapping from $I^i_t$  to 
   a (random) action  $\tilde a_t^i =  (a_t^i, \{ m_{t}^{i\rightarrow j} \colon   i \in \cN_t^j \})$, i.e., $\tilde a_t^i \sim \pi_t^i ( \cdot \given I_t^i)$. 
   Notice that the 
   size of information  set  $I_t^i$ grows as $t$ grows. To handle the memory issue, it is common to first embed $I_t^i$ in a fixed latent space via recurrent neural network (RNN)  or Long Short-Term Memory (LSTM)  \cite{hochreiter1997long} and define the value  and policy functions on top of the embedded features. Moreover, most existing works in this line of research adopt the paradigm of centralized learning and utilize  techniques such as weight-sharing or attention mechanism \cite{vaswani2017attention} to increase computational efficiency.  With centralized learning, single-agent RL algorithms such as Q-learning and actor-critic are readily applicable.    
   
 In particular, 
    \cite{foerster2016learning} first proposes to tackle the problem of learning to communicate via deep Q-learning. They propose  to use two Q networks that govern taking action $a^i \in \cA$ and producing messages separately. Their training algorithm is an extension of the deep recurrent Q-learning (DRQN) \cite{hausknecht2015deep}, which combines RNN and deep Q-learning \cite{mnih2015human}. Following  \cite{foerster2016learning}, various works  \cite{jorge2016learning,sukhbaatar2016learning, havrylov2017emergence,  das2017learning,  peng2017multiagent, mordatch2018emergence, jiang2018learning, jiang2018graph, celikyilmaz2018deep, das2018tarmac, lazaridou2018emergence, cogswell2019emergence} have proposed a variety of neural network architectures to foster communication among agents. 
 These works combine single-agent RL methods with novel  developments in deep learning, and demonstrate their performance  via empirical studies. 
Among these works, \cite{das2017learning, havrylov2017emergence, mordatch2018emergence,lazaridou2018emergence,cogswell2019emergence} have reported the emergence of computational communication protocols among the agents when the RL algorithm is trained from scratch with text or image inputs.  We remark that the algorithms used in these works are more akin to single-agent RL due to centralized learning. For more details overviews of  multi-agent  communication, we  refer the interested readers to Section 6 of  \cite{oroojlooyjadid2019review} and Section 3 of \cite{hernandez2018multiagent}.

\subsection{Competitive Setting} \label{sec:application_comp}

Regarding the competitive setting, in the following, we highlight the recent applications of MARL to \emph{the game of Go} and \emph{Texas hold'em  poker}, which are  archetypal instances of two-player perfect-information and partial-information extensive-form games, respectively.

\vspace{6pt}
\noindent{\bf The Game of Go}
\vspace{3pt}

The game of Go is a board game played by two competing players, with the goal  of  surrounding more territory on the board than the opponent.  These two players have access to white or black stones respectively, and take turns placing their stones on a $19\times 19$ board, representing their territories. In each move, a player  can place a stone to any of the total $361$ positions on the board that is not already taken by a stone. Once placed on the board, the stones cannot be moved. But the stones will be removed from the board when completely surrounded by opposing stones. The game terminates when neither of the players is unwilling or unable to make a further move, and the winner is determined by counting the  area of the territory and the number of stones captured by the players.  

The game of Go can be viewed as a two-player zero-sum Markov game with deterministic state transitions, and the reward  only appears at the end of the game.  The state of this Markov game is the current configuration of the board and the reward is either one or minus one, representing either a win or a loss, respectively. Specifically,  we have $  r^1 (s)+ r^2(s)   =   0$ for any state $s\in \cS$, and   $r^1(s), r^2 (s) \in \{1, -1\}$ when $s$ is a terminating state, and $r^1 (s) = r^2 (s) = 0$ otherwise.  Let $V_{*}^i(s)$ denote the optimal value function of player $i \in \{1,2\}$.  Thus, in this case,  $[ 1+ V^{i}(s)]/2$ is  the probability of player $i\in \{1,2\}$  winning  the game when the current state is $s$ and both players follow the Nash equilibrium policies thereafter.  Moreover,
as this Markov game is turn-based, it is known that the Nash equilibrium policies of the two players are deterministic \cite{hansen2013strategy}. 
Furthermore,  since each configuration of the board can be constructed from a sequence of moves of the two players due to deterministic transitions, we can also view the game of Go as an extensive-form game with perfect information. 
This problem is notoriously challenging due to the gigantic state space. It is estimated in 
 \cite{allis1994searching}  that the size of state space  exceeds $10^{360}$, which forbids the usage of  any traditional reinforcement learning or searching algorithms.
 
A significant breakthrough has been made by the  \emph{AlphaGo}  introduced in  \cite{silver2016mastering},  which is the first computer Go program that defeats a human professional player on a full-sized board. AlphaGo integrates a variety of ideas from deep learning and reinforcement learning, and tackles the challenge of huge state space by representing the policy and value functions using deep convolutional neural networks (CNN) \cite{krizhevsky2012imagenet}. Specifically, both the policy and value networks are $13$-layer CNNs with the same architecture, and a board configuration is represented by $48$ features.  Thus, both the policy and value networks take  inputs of size 
 $19\times 19 \times 48$. 
These two networks are trained through  a novel combination of supervised learning from human expert data and reinforcement learning from Monte-Carlo tree search (MCTS) and self-play. Specifically, in the first stage, the policy network is trained by supervised learning to predict the actions made by the  human players, where the dataset consists of $30$ million positions from the KGS Go server. 
That is, for any state-action pair $(s,a)$ in the dataset, the action $a$ is treated as the response variable and the state $s$ is regarded as the covariate. The weights of  
  the policy network is trained via stochastic gradient ascent to maximize the likelihood function. 
  After initializing the policy network via supervised learning, 
 in the second stage of the pipeline, both the policy and value networks are trained via reinforcement learning and self-play. In particular, new data are generated by  games played between the current policy network and a random previous iteration of the policy network. Moreover, 
 the policy network is updated following policy gradient, 
 and the value network aims to find the value function associated with the policy network and is updated by minimizing the mean-squared prediction error. 
 Finally, when playing the game, the current iterates of the  policy and value networks are combined to produce an improved policy by lookahead search via MCTS. The actual action taken by AlphaGo is determined by such an MCTS policy. Moreover, to improve computational efficiency, AlphaGo uses an asynchronous and distributed version of MCTS to speed up   simulation. 
 
 Since the advent of AlphaGo, an improved version, known as  AlphaGo Zero, has been proposed in
  \cite{silver2017mastering}. 
  Compared with the vanilla AlphaGo, AlphaGo Zero does not use supervised learning to initialize the policy network. Instead, both the policy and value networks are trained from scratch solely via reinforcement learning and self-play. Besides, instead of having separate policy and value functions share the same network architecture, in AlphaGo Zero, these two networks are aggregated into a single neural network structure. Specifically, the policy and value functions  are represented by $(p(s), V(s)) = f_{\theta}(s)$, where $s \in \cS$ is the state which represents the current board, $f_{\theta}$ is a deep CNN with parameter $\theta$, $V(s)$ is a scalar that  corresponds to the value function, and  $p(s)$ is a vector which represents the policy, i.e., for each entry $a \in \cA$, $p_a(s) $ is the probability of taking action $a$ at state $s$. Thus, under such a network structure, the policy and value networks automatically share the same low-level representations of the states. Moreover, the parameter $\theta$ of network $f_{\theta}$ is trained via self-play and MCTS. Specifically, at each time-step $t$, based on the policy $p$ and value $V$ given by $f_{\theta_t}$, an MCTS policy $\pi_t$ can be obtained and a move is executed following policy $\pi_t(s_t)$. Such a simulation procedure continues  until the current  game terminates. Then the outcome of the $t$-th time-step, $z_t \in \{1, -1\}$,  is recorded, according to the perspective of the player at time-step $t$. Then the parameter $\theta$ is updated by following a stochastic gradient step on a loss function $\ell_t$, which is defined as 
  \$
  \ell_t(\theta) = [ z - V(s_t) ] ^2 - \pi_t^\top \log p(s_t) + c\cdot \|\theta \|_2^2 , \qquad \big ( p(\cdot) , V(\cdot)  \big) = f_{\theta} (\cdot) . 
  \$
  Thus, $\ell_t$ is the  sum of the mean-squared prediction error of the value function, cross-entropy loss between the policy network and the MCTS policy, and a weight-decay term for regularization. It is reported that 
  AlphaGo Zero has defeated the strongest versions of the previous AlphaGo and that it   also has demonstrated non-standard Go strategies  that had not been discovered  before. Finally, the techniques adopted in AlphaGo Zero has been generalized to other challenging board games. Specifically, 
\cite{silver2018general} proposes the AlphaZero program that is trained by self-play and reinforcement learning with zero human knowledge, and  achieves superhuman performance in the  games of chess, shogi, and Go. 

\vspace{6pt}
\noindent{\bf Texas Hold'em Poker}
\vspace{3pt}

Another remarkable applicational achievement of MARL in the competitive setting focuses on developing artificial intelligence in the Texas hold'em poker, which is one of the most popular variations of the poker. Texas hold'em is usually played by a group of two or more players,
where each player is first dealt with two  \emph{private cards} face down. Then five \emph{community cards} are dealt face up in three  rounds. 
In each round. each player has four possible actions -- \emph{check}, \emph{call}, \emph{raise}, and \emph{fold}. 
After all the cards are dealt, each player who has not folded have seven cards in total, consisting of five community cards and two private cards. Each of these players  then  finds the best five-card poker hand out of all combinations of the seven cards. The player with the best hand is the winner and wins all the money that the players wager for that hand, which is also known as the \emph{pot}. Note that  each hand of Texas hold'em  terminates after three rounds, and the payoffs of the  player are only known after the hand ends. Also notice that each player is unaware of the private cards of the rest of the players. Thus, Texas hold'em is an instance of  multi-player extensive-form game  with incomplete information. 
The game is called \emph{heads-up} when there are only   two
players. When both the bet sizes and the amount of allowed raises are fixed, the game is called \emph{limit hold'em}. In the no-limit hold'em, however, each player may bet or raise any amount  up to all of the money the player has at the table, as long as  it exceeds the previous bet or raise. 

There has been quest for developing superhuman computer poker programs for over two decades \cite{billings2002challenge, rubin2011computer}. Various methods have been shown successful for simple variations of poker such as Kuhn  poker   \cite{kuhn1950simplified} and Leduc hold'em \cite{southey2005bayes}. However, the full-fledged Texas hold'em  is much more challenging and several breakthroughs have been achieved only recently.  The simplest version of Texas hold'em  is \emph{heads-up limit hold'em} (HULHE), which  has $3.9 \times 10^{14}$ information sets in total \cite{bowling2015heads}, where a player is required to take an action at each information set. \cite{bowling2015heads} has for the first time reported  solving HULHE to approximate Nash equilibrium  via  $\text{CFR}^{+}$ \cite{tammelin2014solving,tammelin2015solving}, a variant of counterfactual regret minimization \cite{zinkevich2008regret}. Subsequently, other methods such as Neural Fictitious Self-Play  \cite{heinrich2016deep}  and Monte-Carlo tree search with self-play \cite{heinrich2015smooth} have also been adopted to successfully solve HULHE. 

Despite these   breakthroughs, solving  \emph{heads-up no-limit hold'em} (HUNL) with artificial intelligence has remained open until recently, which has more than $6\times 10^{161}$ information sets, an astronomical number. Thus, in HUNL,  it is impossible (in today's computational power)  to traverse  all information sets, making it inviable to  apply  $\text{CFR}^{+}$ as in \cite{bowling2015heads}. Ground-breaking achievements have   recently been made by  \emph{DeepStack}  \cite{moravvcik2017deepstack} and \emph{Libratus}  \cite{brown2018superhuman}, two computer poker programs developed independently,  which defeat human professional poker players in HUNL for the first time. Both of these  programs adopt CFR as the backbone  of their algorithmic frameworks,  but  adopt different strategies for handling the gigantic size of the game. In particular, DeepStack applies  deep learning to learn good representations of the game and proposes \emph{deep counterfactual value networks} to integrate deep learning and CFR. Moreover, DeepStack adopts 
  limited depth lookahead planning to reduce the gigantic $6 \times 10^{161}$ information sets to no more than $10^{17}$ information sets, thus making it possible to enumerate all information sets.  
  In contrast, \emph{Libratus}   does not utilize any  deep learning techniques. Instead, it reduces the size of the game by computation of an abstraction of the game, which is possible since many of the information sets are very similar.  
  Moreover, it further reduces the complexity by using the  sub-game decomposition technique \cite{burch2014solving, moravcik2016refining, brown2017safe} for imperfect-information games and by constructing fine-grained abstractions  of the sub-games. When the abstractions are constructed, an improved version of the Monte-Carlo CFR \cite{lanctot2009monte, burch2012efficient,gibson2012generalized} is utilized to compute the policy. 
 Furthermore, very recently, based upon \emph{Libratus}, \cite{brown2019superhuman} has proposed  \emph{Pluribus}, a computer poker program  that has been shown to be stronger than top human professionals in   no-limit Texas hold'em poker with six players. The success of Pluribus is attributed to the  following  techniques that have appeared in the literature: abstraction and sub-game decomposition for large-scale imperfect-information games, Monte-Carlo CFR, self-play, and depth-limited search. 

    \vspace{6pt}
\noindent{\bf Other Applications}
\vspace{3pt}

Furthermore, another popular testbed of MARL is the StarCraft II \cite{vinyals2017starcraft}, which is an immensely popular multi-player real-strategy computer  game.  This game can be formulated as a multi-agent Markov game with partial observation, where  each player  has only limited information of the game state. Designing reinforcement learning systems for StarCraft II is extremely challenging due to the needs to make  decisions  under uncertainty and incomplete information, to consider the optimal strategy in the long-run, and to design good reward functions that elicit learning. Since released, both the full-game and sub-game versions of  StarCraft II have gained tremendous research interest. 
A   breakthrough in this game was   achieved by    \emph{AlphaStar},   recently proposed in  \cite{vinyals2019grandmaster}, which has demonstrated superhuman performance in zero-sum two-player full-game StarCraft II.  Its reinforcement learning algorithm  combines   LSTM for the  parametrization of policy and value functions, asynchronous actor-critic  \cite{mnih2016asynchronous} for policy updates, and Neural Fictitious Self-play  \cite{heinrich2016deep} for equilibrium finding. 

\subsection{Mixed Settings}

Compared to the cooperative and competitive settings,  research on MARL  under the mixed setting is rather less explored. One  application in this setting is multi-player poker. As we have mentioned in \S\ref{sec:application_comp},  Pluribus introduced in  \cite{brown2019superhuman}  has  demonstrated superhuman performance in  six-player no-limit Texas hold'em.  In addition, as an extension of the problem of learning to communicate, introduced in 
\S\ref{sec:application_coop}, there is a line of research that aims to  apply MARL to tackle 
learning  social dilemmas, which is usually formulated as a multi-agent stochastic game with partial information. Thus, most of the algorithms proposed under these settings incorporate RNN or LSTM for learning representations of the histories experienced by the agent, and the performance  of  these algorithms are usually exhibited using experimental results;  see, e.g., \cite{leibo2017multi, lerer2017maintaining, hughes2018inequity}, and the references therein.

Moreover, another example of the mixed setting is the case where the agents are divided into two opposing teams that play zero-sum games. The reward of a team is shared by each player within this team. Compared with two-player zero-sum games, this setting is more challenging in that both cooperation among teammates and competition against the opposing team  need to be taken into consideration. A   prominent testbed of this case is the \emph{Dota 2} video game, where  each of the two teams, each with five players, aims to conquer the base of the other team and defend its own base. Each player independently controls a powerful character known as the \emph{hero}, and only observes the state of the game via the video output on the screen. Thus, Dota 2 is a zero-sum Markov game  played by two teams, with each agent having imperfect information of the game. For this challenging problem, in 2018, \emph{OpenAI} has proposed the \emph{OpenAI Five} AI system \cite{OpenAI_dota}, which enjoys  superhuman performance and has  defeated human world champions in an e-sports game. The algorithmic framework integrates LSTM for learning good representations and proximal policy optimization \cite{schulman2017proximal} with self-play for policy learning. Moreover, to balance between effective coordination and communication cost, instead of having 
explicit communication channels among the teams, OpenAI Five utilizes reward shaping by having a hyperparameter, named ``team spirit'', to balance the relative importance between each hero's individual reward function  and the average of the team's reward function.

%


\section{Conclusions and  Future  Directions}\label{sec:conclusion}

Multi-agent RL has long been an active and significant research area in reinforcement learning, in view of the  ubiquity of  sequential decision-making with multiple agents coupled in their actions and information. In stark contrast to its great empirical success, theoretical understanding of MARL algorithms is well recognized  to be challenging and relatively lacking in the literature. Indeed,  establishing an encompassing theory for MARL requires tools spanning  dynamic programming, game theory, optimization theory, and statistics, which are  non-trivial  to unify and investigate within one context. 

In this chapter, we have provided a selective overview of mostly recent MARL algorithms, backed by   theoretical analysis, followed by several high-profile but challenging applications that have been addressed lately. Following  the classical overview \cite{bu2008comprehensive}, we have categorized the algorithms into three groups: those solving problems that are  fully cooperative, fully competitive, and a mix of the two.  Orthogonal to the existing reviews on MARL, this chapter has laid emphasis on several new angles and taxonomies of MARL theory,  some of which have been drawn  from our own research endeavors and interests. We note that our overview  should not be viewed  as a  comprehensive one, but instead as a focused one dictated by our own interests and expertise,  which should appeal to  researchers of similar interests, and provide  a stimulus for  future research directions in this general topical area.  Accordingly, we have identified 
 the following paramount while open  avenues for future research on MARL theory.   
 
\vspace{6pt}
\noindent{\bf Partially observed settings:} 
\vspace{3pt} 
Partial observability of the system states and the actions of other agents is quintessential and inevitable in many practical MARL applications. In general, these settings can be modeled as a partially observed stochastic game (POSG), which includes the cooperative setting with a common reward function, i.e., the Dec-POMDP model, as a special case. Nevertheless, as pointed out in \S\ref{subsec:partially_observed}, even the cooperative task is NEXP-complete  \cite{bernstein2002complexity} and difficult to solve. In fact, the information state for optimal decision-making in POSGs can be  very complicated  and involve  belief generation over the opponents' policies 
\cite{hansen2004dynamic}, compared to that  in POMDPs, which requires belief on only states. This difficulty essentially stems  from the heterogenous beliefs of agents  resulting from their own observations obtained from the model, 
an inherent challenge of MARL  mentioned  in \S\ref{sec:challenges} due to various information structures. It might be possible to start by  generalizing the centralized-learning-decentralized-execution scheme for solving Dec-POMDPs \cite{amato2015scalable,dibangoye2018learning} to solving POSGs. 

\vspace{6pt}
\noindent{\bf Deep MARL theory:} 
\vspace{3pt} 
As mentioned in \S\ref{subsec:scala_issue}, using deep neural networks for function approximation can address the scalability issue in MARL. In fact, most of the recent empirical successes in MARL result from the use of DNNs \cite{heinrich2016deep,lowe2017multi,foerster2017counterfactual,gupta2017cooperative,omidshafiei2017deep}. Nonetheless, because of  lack of theoretical backings, we have not included details of these algorithms in this chapter. Very recently, a few attempts have been made to understand the global convergence of several single-agent deep RL algorithms, such as  neural TD learning \cite{cai2019neural} and  neural policy optimization  \cite{wang2019neural,liu2019neural},  when overparameterized neural networks \cite{arora2018optimization,li2018learning} are used. It is thus promising to extend these results to multi-agent  settings, as initial steps toward theoretical understanding of deep MARL. 

\vspace{6pt}
\noindent{\bf Model-based MARL:} 
\vspace{3pt} 
It may be slightly surprising that very few MARL algorithms in the literature are \emph{model-based}, in the sense that   the MARL model is first estimated, and then used as a nominal one to design algorithms. To the best of our knowledge, 
the only existing model-based MARL  algorithms include the early one in \cite{brafman2000near} that solves single-controller-stochastic games, a special zero-sum MG; and the later improved one in \cite{brafman2002r}, named R-MAX, for zero-sum MGs. These  algorithms are also built upon the principle of {optimism in the  face of uncertainty} \cite{auer2007logarithmic, jaksch2010near}, as several aforementioned model-free ones. Considering recent progresses in model-based RL, especially its provable advantages over model-free ones in certain regimes  \cite{tu2018gap,sun2019model}, it is worth generalizing these results to MARL to improve its sample efficiency. 
 
\vspace{6pt}
\noindent{\bf Convergence of policy gradient methods:}
\vspace{3pt}
As mentioned in  \S\ref{subsec:mixed_setting}, the convergence result of vanilla policy gradient method in general MARL is  mostly negative, i.e., it may avoid even the local NE points in many cases.   
This is essentially related to the challenge of non-stationarity in MARL, see \S\ref{subsec:non_stationary}. Even though some remedies have been advocated  \cite{balduzzi2018mechanics,letcher2019differentiable,chasnov2019convergence,fiez2019convergence} to stabilize the convergence in \emph{general continuous} games, these assumptions are not easily verified/satisfied in MARL, e.g., even in the simplest LQ setting \cite{mazumdar2019policy},   as they depend not only on the model, but also on the policy parameterization. Due to this subtlety, it may be interesting to explore the (global) convergence of policy-based methods for MARL, probably starting with the simple LQ setting, i.e., general-sum LQ games, in analogy to that for the zero-sum counterpart \cite{zhang2019policyb,bu2019global}. 
Such an exploration may also benefit from the recent advances of nonconvex-(non)concave optimization  \cite{lin2018solving,jin2019minmax,nouiehed2019solving}. 

\vspace{6pt}
\noindent{\bf MARL with robustness/safety concerns:} 
\vspace{3pt}
Concerning the challenge of non-unique learning goals in MARL (see \S\ref{subsec:learning_goal}), we believe it is of merit to consider robustness and/or safety constraints in MARL. To the best of our knowledge, this is still a relatively uncharted territory. In fact, safe RL has been recognized as one of the most significant challenges  in the single-agent setting \cite{garcia2015comprehensive}. With more than one agents that may have conflicted objectives, guaranteeing  safety   becomes more involved, as the safety requirement now concerns the coupling of all  agents. One straightforward model is constrained multi-agent MDPs/Markov games, with the constraints characterizing the safety requirement. Learning with provably safety guarantees in this setting is non-trivial, but  necessary for some safety-critical MARL applications as autonomous driving \cite{shalev2016safe} and robotics \cite{kober2013reinforcement}. In addition, it is also natural to think of  robustness against adversarial agents, especially in the decentralized/distributed cooperative MARL settings as in \cite{zhang2018fully,chen2018communication,wai2018multi}, where the adversary may disturb the learning process in an anonymous way -- a common scenario  in distributed systems. Recent development of robust distributed supervised-learning against Byzantine adversaries  \cite{chen2017distributed,yin2018byzantine} may be useful  in this context.



{\small
\bibliographystyle{spmpsci_new}  
\bibliography{MARL_Springer_1,MARL_Springer_2}}

\end{document}